%% file: main.tex
\documentclass{article}

\usepackage{iclr2025_conference}
\usepackage{times}

\usepackage[utf8]{inputenc} 
\usepackage[T1]{fontenc}    
\usepackage{hyperref}       
\usepackage{url}            
\usepackage{booktabs}       
\usepackage{amsfonts}       
\usepackage{nicefrac}       
\usepackage{microtype}      
\usepackage{xcolor}         
\usepackage{ragged2e}

\usepackage{bbm}
\usepackage{graphics}
\usepackage{graphicx}
\usepackage{booktabs}
\usepackage{enumitem}
\usepackage{algorithm}
\usepackage{algorithmic}
\usepackage{arydshln}
\usepackage{multirow}
\usepackage{tablefootnote}
\usepackage{threeparttable}
\usepackage{subcaption}
\usepackage{xspace}
\usepackage{wrapfig}
\usepackage{lipsum}
\usepackage{tabularx}
\usepackage{pifont}
\usepackage{array}

\usepackage{hyperref}
\usepackage{amsmath}
\usepackage{makecell}

\input{preamble}
\newcommand{\cmark}{\ding{51}}%
\newcommand{\xmark}{\ding{55}}%
\definecolor{mygreen}{HTML}{336633}
\definecolor{myred}{HTML}{CC3300}

\makeatletter
\DeclareRobustCommand\onedot{\futurelet\@let@token\@onedot}
\def\@onedot{\ifx\@let@token.\else.\null\fi\xspace}

\def\eg{\emph{e.g}\onedot} 
\def\ie{\emph{i.e}\onedot}

\makeatother
\iclrfinalcopy

\def\dataset{Shot2Story }
\title{Shot2Story: A New Benchmark for Comprehensive Understanding of Multi-shot Videos}

\author{%
  Mingfei Han$^{2,3,5\dagger}$\thanks{Work was done during an internship at Bytedance. $^{\dagger}$ Equal contribution.}  ,
  Linjie Yang$^{1\dagger}$, 
  Xiaojun Chang$^{3,4}$, 
  Lina Yao$^{5}$, 
  Heng Wang$^{1}$ \\
  $^{1}$Bytedance Inc.\quad
  $^{2}$ReLER Lab, AAII, UTS\quad
  $^{3}$Department of Computer Vision, MBZUAI\quad\\
  $^{4}$University of Science and Technology of China\quad
  $^{5}$Data61, CSIRO\\
  \url{https://github.com/bytedance/Shot2Story}
}

\begin{document}

\maketitle

\input{sec/0_abstract}  
\input{sec/1_intro}
\input{sec/dataset}
\input{sec/experiments}
\input{sec/conclusion}

\newpage
\bibliography{main}
\bibliographystyle{iclr2025_conference}

\appendix
\input{sec/X_suppl}

\end{document}

%% file: preamble.tex
\usepackage{colortbl}
\newcommand{\videocnt}{42,958\xspace}
\usepackage{caption}

%% file: sec/0_abstract.tex
\begin{abstract}
A short clip of video may contain progression of multiple events and an interesting story line. A human need to capture both the event in every shot and associate them together to understand the story behind it. In this work, we present a new multi-shot video understanding benchmark \dataset with detailed shot-level captions, comprehensive video summaries and question-answering pairs. To facilitate better semantic understanding of videos, we provide captions for both visual signals and human narrations. We design several distinct tasks including single-shot video captioning, multi-shot video summarization, and multi-shot video question answering. Preliminary experiments show some challenges to generate a long and comprehensive video summary for multi-shot videos. Nevertheless, the generated imperfect summaries can already 
achieve competitive performance on existing video understanding tasks such as video question-answering, promoting an under-explored setting of video understanding with detailed summaries.
\end{abstract}


%% file: sec/1_intro.tex
\section{Introduction}
\label{sec:intro}


Video captioning is a long-standing video understanding task to facilitate open-world video analysis with the help of human-annotated captions. Since a video may contain multiple events, dense captioning benchmarks (Ego4D~\citep{grauman2022ego4d}, YouCook2~\citep{zhou2018youcook2}, ActivityNetCaps~\citep{krishna2017activitynet}) capture information of multiple events in videos ranging from 3-20 minutes. However, even within seconds,  more than one event often occurs in daily videos such as news broadcasts, tutorial videos, and movies. Specifically, shot transition, which is a common technique to transfer from one event to another, or to switch the viewpoint of a single event, happens less than every 4s for average English movies after 2010~\citep{cutting2011shot}. Although some existing captioning benchmarks~\citep{xu2016msr-vtt, krishna2017activitynet,zhou2018youcook2} already use multi-shot videos, they often annotate the captions in a coarse-grained manner, either providing a holistic caption or asking annotators to subjectively choose the boundary of each event. To better accommodate the multi-shot formation of videos, we believe a new video benchmark with rich textual descriptions based on video shots is favored in the research community.

\begin{figure}
\centering
\includegraphics[width=\linewidth]{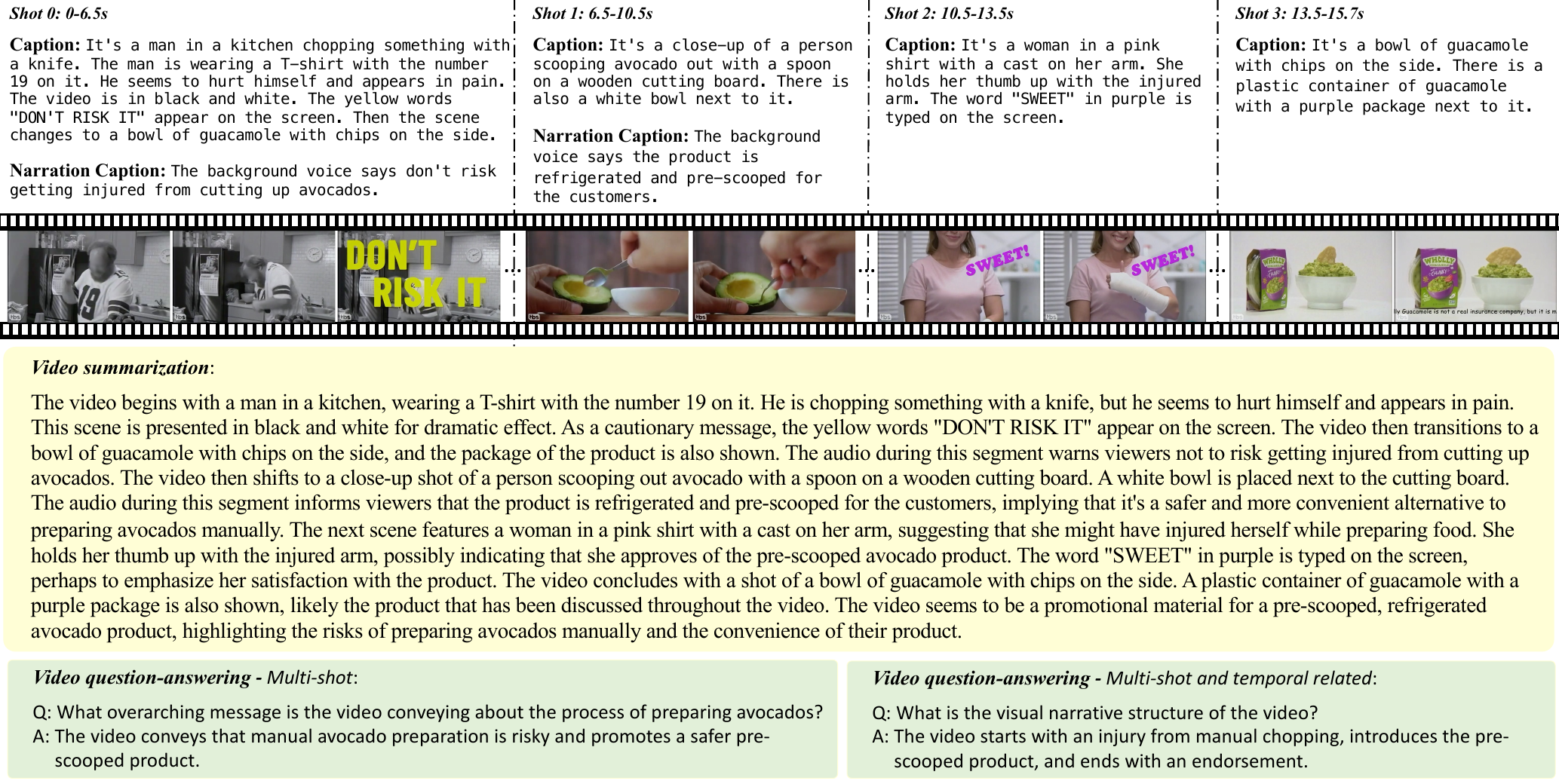} 
\vspace{-4mm}
\caption{An annotated example of our \dataset with sing-shot visual captions and narration captions. Moreover, we provide coherent and reasonable video summaries, and question-answering pairs to facilitate comprehensive understanding of multi-shot videos.}
\label{fig:video_demo}
\vspace{-3mm}
\end{figure}

On the other hand, multi-shot videos are often accompanied by rich narrations that relate to the different events happening in the video. A model needs to capture both the visual and audio signals to understand the underlying story. Specifically, narrations may contain key information that cannot be inferred from pure visual information only. See Figure~\ref{fig:video_demo}, without the narration, a viewer is unable to capture the relationship between the man's action and the avocado product in the first shot. 

In this work, we propose a new benchmark \dataset for audio-visual understanding of multi-shot videos. We collect a dataset of \videocnt short videos where the average number of shots in each video is 4.4. For each video shot, we annotate a detailed textual description for the video frames and another textual description for the human speech. We also leverage a state-of-the-art large language model (LLM) GPT-4~\citep{chatgpt} to generate a long textual video summary from the annotated clip descriptions, which are further verified by human annotators. The summary includes additional details such as transitions of different shots, progression of multiple events, and mapping of the subject identities in different scenes. An example of one annotated video is shown in Figure~\ref{fig:video_demo}.

To benchmark the advances of multi-modal video understanding, we designed several distinctive tasks using our dataset, including single-shot video captioning, multi-shot video summarization, and multi-shot video question answering. We design and implement several baseline models using a frozen vision encoder and an LLM, by prompting the LLM with frame tokens and ASR (Automatic Speech Recognition) text. Through extensive experiments, we show that: 
(1) the ASR text is critical to joint understanding of visual and audio content, 
(2) processing the video as a whole without the shot structure degenerates the model's capacity of understanding the multi-shot video, 
(3) the summarization model trained on our multi-shot summaries can be used on the proposed multi-shot QA benchmark and generalized to other datasets with longer durations (ActivityNet\citep{krishna2017activitynet}) and out-of-domain topics (MSRVTT\citep{xu2016msr-vtt}), validating the quality of our annotated summaries. 
Without any bells and whistles,  we attain competitive results on zero-shot video question-answering by converting the problem into pure text-based QA with the generated video summaries.


%

%% file: sec/dataset.tex
\section{The \dataset benchmark}
\label{sec:dataset}

\begin{figure*}[!htb]
  \centering
  \includegraphics[width=1.\linewidth]{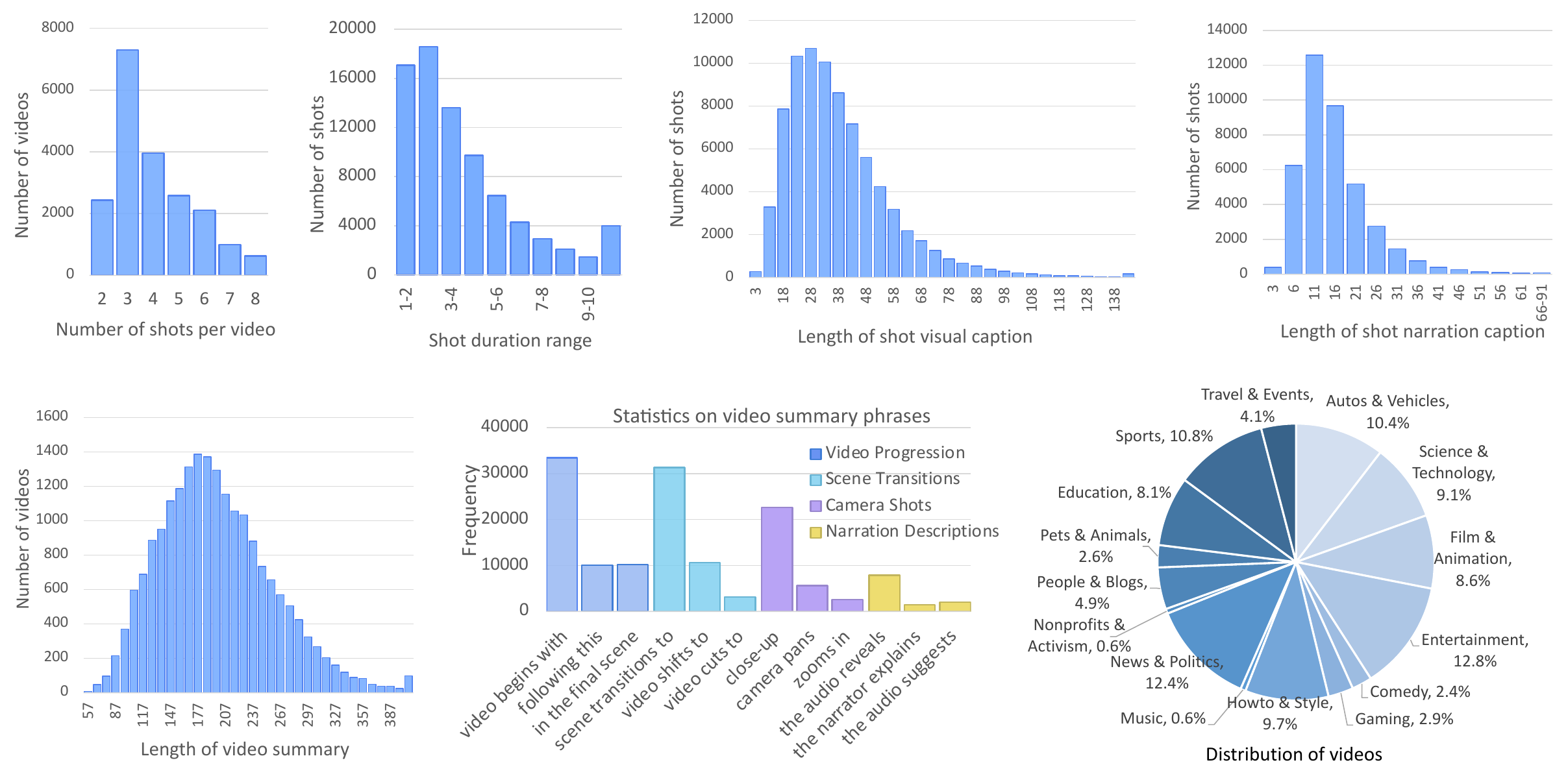}
  \vspace{-5mm}
  \caption{Statistics of \dataset. Our dataset features detailed visual captions and narration captions, and video summaries, highlighting video progressions, transitions, camera cuts and narration descriptions, with statistics of frequent expressions depicted in the figure.}
  \vspace{-4mm}
  \label{fig:stats_overview}
\end{figure*}

Our new benchmark \dataset contains \videocnt videos. The length of each video is ranging from 10s to 40s.
We first use an off-the-shelf shot detection method TransNetV2~\citep{souvcek2020transnet} to split each video into shots. 
For each video shot, we annotate captions for both visual and audio information. Then we further annotate video summaries based on the annotated shot captions. Figure~\ref{fig:stats_overview} shows an overview of our dataset with some key statistics.

\vspace{-2mm}
\subsection{Data preparation}
\label{sec:data_preparation}
\vspace{-2mm}

We source videos for our dataset from the public video benchmark HDvila100M~\citep{xue2022hdvila}. It offers a large collection of narrative videos, comprising 3M YouTube videos segmented into 100M clips, each about 13 seconds long.
We choose this data source for its concise yet complex multi-shot formats, diverse topics, and abundant ASR content. Since we prefer videos with both rich visual and ASR information, we design several filtering techniques to exclude those videos with either low visual-ASR correlation or static visual content.

We start with keeping video clips with durations between 10 to 40 seconds, since we observe that the majority of the video clips from HDvila100M fall in this range. 
Then we remove videos with more than 8 shots due to the heavy annotation cost. We also notice that the video segments with too many shots in HDvila100M tend to be slideshows or image collages that deviates from our focuses. 
Further, to harvest videos with rich visual-ASR correlations, we set up a metric between video shots and ASR texts. Specifically, we uniformly sample 4 video frames for each shot and obtain the cosine similarity score between the video shot embedding and the text embedding using CLIP~\citep{radford2021learning} encoders. We only keep the videos containing at least one shot that is visually correlated to ASR with a threshold of 0.25.
Next, in order to obtain videos with diverse shot contents, we set up an inter-shot metric to filter out the videos with similar adjacent shots. We compute the cosine similarities between embeddings of adjacent shots and keep the videos with all inter-shot similarity scores smaller than 0.9.
Finally, to further remove the videos with static contents, we adopt an intensity-based scene change detector in PySceneDetect\footnote{https://www.scenedetect.com/}
 with a low threshold of 11 on our segmented shots. If the filter is unable to detect scene changes at this low threshold, it is conceivable that the shot contains static contents. We only keep the video clips in which all shots contain no static content based on our filtering method.

As a result, from a total of 2.1M sampled video clips from HD-VILA-100M, we obtain $42,958$ video clips that meet our quality standard. The number of shots in each video is from 2 to 8.
These videos are then shared with our annotators for further annotations. 

\vspace{-2mm}
\subsection{Annotation of single-shot captions}
\label{sec:generate_caption}
\vspace{-2mm}
After using TransNetV2 to divide the target videos into video shots, we ask annotators to 
annotate both visual-only captions and audio-related captions for each shot. 
We split these two caption annotation
to facilitate separate modeling of these two types of information source.
For visual-only captions, we require annotators to describe the major subjects and events in the video. Since it is an open-world setting, the videos can be quite diverse and hard to describe. In order to reduce the difficulties of annotating a caption from scratch, we generate an initial video caption using MiniGPT-4~\citep{zhu2023minigpt} by sampling 4 image frames from the video clip and prompting the model using below prompt.


\emph{\#\#\#Human:\textless Img\textgreater Frame1\textless/Img \textgreater \textless Img\textgreater Frame2\textless/Img\textgreater \textless Img\textgreater Frame3\textless/Img\textgreater \textless Img\textgreater Frame4\textless/Im \\
g\textgreater Please describe this video. Do not include details that you are not sure of. For example, if there is text in the image, do not include the content of the text if they are not clearly shown. \#\#\#Assistant: } 

Although MiniGPT-4 is originally designed for image understanding, empirically it is able to generate captions for short video clips, both comprehensively and reasonably. It is able to describe different subjects including persons, animals, food, tools, and virtual objects like animated characters. 
Annotators are first instructed to correct any errors in the original captions. The mistakes include incorrect descriptions of the object categories, attributes, actions, facial expressions, etc. Also, there might be some subjective descriptions generated by MiniGPT-4 such as emotion and atmosphere. We ask annotators to remove all these subjective descriptions. 
We then ask annotators to supplement the information about the major subjects, actions, and backgrounds present in the video. The goal is for the resulting captions to accurately capture the key elements of each video shot. Statistics shows over 80\% of single-shot visual captions are manually corrected. For narration captions, annotators should watch the video and interpret the audio content that visually correlates into descriptive narration, including information sources. For example: “According to the woman in white, the room is not very clean.” All narration captions are manually drafted from scratch.
An example of this annotation process is shown in Appendix~\ref{supp_sec:human_anno_shot}, where the annotator corrects the caption from ``standing in front of the car'' to ``getting close to the car'', and adding a missing detail of ``a close-up shot of the front''. In this way, we find the annotation speed significantly faster ($\small \sim$ $3 \times$) compared to writing a caption from scratch. On the other hand, we find the captions generated this way has more coherent style and tend to cover more details of the video.

\begin{table}[!t]
\caption{High-level comparison of our dataset to previous ones. The summary length of ActivityNet and YouCook2 are combined length of captions in one video. M and G denote manual and generated.}
\vspace{-2mm}
\centering
\small
\setlength{\tabcolsep}{2.5pt} 
\resizebox{1\textwidth}{!}{
\begin{tabular}{@{}l c c c c c l l l@{}}
\toprule
Dataset          & Annotation      & \thead{Multi-shot\\ Video} & \thead{Multi-event  \\Descriptions
} & \thead{Detailed\\ Summary} &  \thead{Summary\\ Length} &  \#Videos & \thead{Avg.\\ Duration} \\ 
\midrule
MSRVTT~\citep{xu2016msr-vtt}           & M           & \cmark     & \xmark                       & \xmark    & -    & 10K     & 15s      \\                                 
ActivityNet Caps~\citep{krishna2017activitynet} & M           & \cmark   & \cmark                & \xmark  & 52.4 & 20K     & 3min                                     \\ 
VideoStorytelling~\citep{li2019video}            & M           & \cmark   & \cmark         & \cmark &  162.6     &  105  & 12.5min                                     \\ 
Ego4D~\citep{grauman2022ego4d}            & M           & \xmark   & \cmark                 & \xmark & -     & 10K     & 23min                                     \\ 
YouCook2~\citep{zhou2018youcook2}         & M           & \cmark   & \cmark                 & \xmark & 67.8     & 2K      & 6min                                      \\ 
VAST~\citep{chen2024vast}             & G        & \cmark   & \xmark                 & \cmark  &  32.4  & 27M     & 5$\sim$30s                           \\ 
\midrule
\dataset             & M+G & \cmark   & \cmark               & \cmark   & 218.3  & 43K & 17.1s \\
\bottomrule
\end{tabular}
}
\vspace{-4mm}
\label{tab:compare}
\end{table}


In contrast to the traditional video captioning benchmarks~\citep{xu2016msr-vtt,krishna2017activitynet,zhou2018youcook2}, we also annotate narration captions in addition to the visual-only captions. Different from existing audio captioning benchmarks~\citep{gemmeke2017audioset}, we focus more on human speeches rather than acoustic events. Annotators are required to associate the human speech with the video content and summarize the main idea of the speech. We require annotators to describe the source of the speech using visual information. For example, if someone is talking, the annotators need to describe which person in the video is talking. If the human speech refers to some object in the video, the annotator is required to describe which object in the video the speaker is referring to. Note that the speaker identity and reference of visual objects are critical information for understanding a video that cannot be trivially obtained using existing algorithms. There are existing research on speaker identification~\citep{kim2021speakerdet} and visual grounding~\citep{anne2017localizing,zhou2019grounded}, but they only work well on constraint scenarios. 

\vspace{-2mm}
\subsection{Annotation of video summary}
\vspace{-2mm}
\label{sec:anno_video_sum}

To create video summaries with annotated video-shot captions, we leverage an LLM-based approach. Specifically, we form a text prompt with incorporating all shot captions and ASR text included, and uses GPT-4~\citep{chatgpt} to generate a cohesive summary. The text prompt we use is shown in Appendix~\ref{supp_sec:gpt4_prompt}. 
The quality is assured through further review and correction by our annotators.

We prompt GPT-4 to produce coherent, fluent text summaries with transition expressions such as ``the video begins'', ``following this'', and ``in the final scene'' to connect video-shot descriptions. The generated annotations also encompass a higher-level understanding of shots, using key phrases such as ``scene shifts back'' and ``returns to the scene'' to denote recurring scenes across shots. Notably, GPT-4 often identifies and links the same subjects across scenes without relying on explicit re-identification models. It draws on descriptive and attributive text from shot captions like ``a newsroom'' or ``a man wearing a black suit'' to infer scene or subject identity. 
To ensure quality, annotators carefully review and correct any inconsistencies in scene or subject references within summaries.
Since our shot-level captions for generating video summaries are manually checked and annotated, the initial video summaries merely have factual errors, with exceptions for some subject identity and scene mismatches. We require annotators to pay more attention to these errors and ensure holistic summary is accurate and comprehensive. Statistics show that over 40\% summaries are manually corrected.

Despite the rigorous verification process, the reliance on automated generation introduces certain inaccuracies and biases. Common pitfalls in the generated summaries include the omission of minor yet contextually important details and a bias towards emphasizing more prominent actions or objects, potentially overlooking less conspicuous elements. This is partly due to our dataset predominantly featuring human-centric activities, as a result of our video filtering process that selects videos rich in visually related audio information and sourcing from HDVILA~\cite{xue2022hdvila}, which primarily curates content related to human activities. Consequently, our annotations tend to highlight salient events and large-scale objects essential to the video’s storyline, mentioning smaller objects only when they directly contribute to the narrative.



\vspace{-2mm}
\subsection{Annotation of question-answering pairs}
\label{sec:qa_anno}
\vspace{-2mm}

We annotate the question-answering pairs on videos in validation and testing splits.
To construct this benchmark, we begin with the human-annotated video summaries from Section~\ref{sec:anno_video_sum}, which is detailed in video content. We then prompt GPT-4~\citep{achiam2023gpt} to generate candidate question-answer pairs in three predefined categories: temporal-related (\eg, \textit{Does the woman appearing at the end of the video wear any accessories?}), multi-shot holistic understanding (\eg, \textit{What is the overarching theme of the video?}), and audio content related (\eg, \textit{Which specific car model does the background voice mention, and what visual features confirm its identity?}). The text prompt we use is shown in Appendix~\ref{supp_sec:prompt_qa}.

The quality is then assessed through further automatic filtering and manual checks. 
Annotators are instructed to verify each question-answer pair against the video content and discard any with mistakes. Simultaneously, they are asked to categorize the questions, where a single question might fall into multiple categories, facilitating evaluating different aspects of multishot understanding \ie, understanding sequences of events and actions (temporal related), integrating information across multiple shots (holistic understanding), correlating audio content with visual elements (audio related), and others (directly discarded). 
After this verification, annotators carefully review the quality of the questions to ensure they solely correlate with the video content and that the answers are not revealed in the question texts. This step includes a thorough manual check to address potential mistakes and biases from the initial verification. Through these sequential annotation stages, we ensure high-quality annotations. 
The process starts with detailed manual single-shot captioning, followed by careful verification and correction at each stage. Even with LLMs introduced to reduce workload, the human-involved process and thorough procedures keep the resulting annotations well-aligned with human labeling.

\begin{wrapfigure}{r}{0.4\textwidth}
  \vspace{-5mm}
  \begin{center}
    \includegraphics[width=0.4\textwidth]{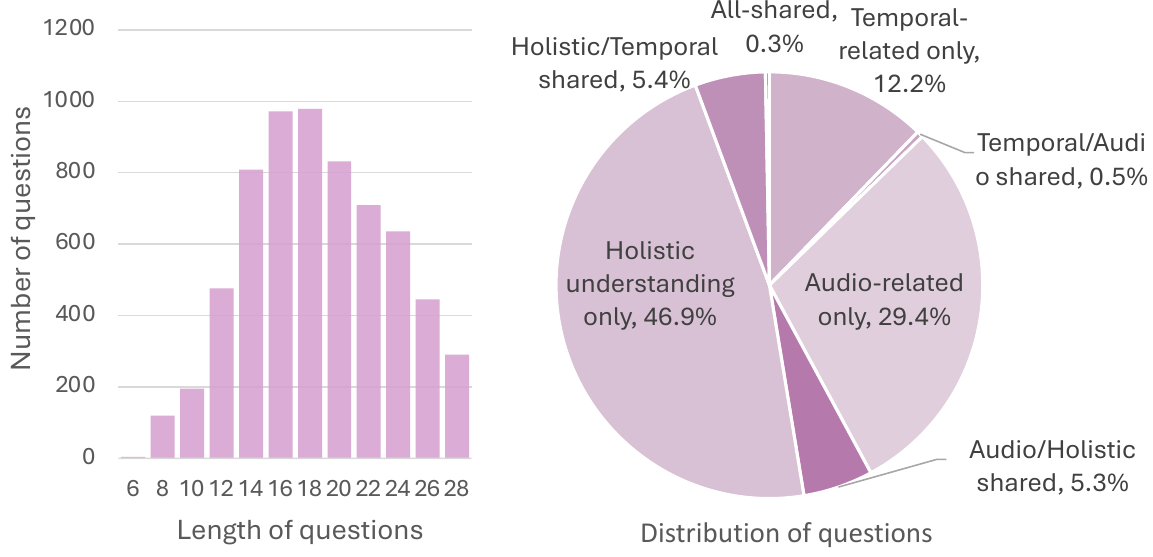}
  \end{center}
  \caption{\footnotesize Distribution of multi-shot video QA benchmark. Questions from different categories overlap. \textit{All-shared} means questions fall under all three categories.}
  \label{fig:qa_stats}
  \vspace{-3mm}
\end{wrapfigure}
Subsequently, to optimize clarity and reduce unnecessary information in the QA pairs, we remove QA pairs with questions exceeding 28 words or answers exceeding 20 words.
Then, we employ Vicuna-13B~\citep{vicuna2023} to attempt answering the questions without video context and discard those answerable without accessing the videos.
Finally, the pairs are tested against popular 
methods such as Video-LLaVA~\citep{lin2023video}, LLaMA-VID~\citep{li2023llama}, Video-ChatGPT~\citep{maaz2023videochatgpt}, and VideoChat2~\citep{li2023mvbench}. 
Questions correctly answered by more than two models are excluded to guarantee that our dataset poses a substantial challenge. Finally, we obtain 4905 QA pairs for validation and 6465 QA pairs for testing set.

\vspace{-3mm}
\subsection{Comparison to existing benchmarks}
\vspace{-1mm}


Compared to existing video description datasets, our dataset is more challenging due to the explicit modeling of the multi-shot nature of web videos. Our textual description includes both shot-level captions and video-level summaries, combining visual and audio understanding, which provides a unique test bed for multi-modal video understanding. Table~\ref{tab:compare} shows a high-level comparison of our new dataset with existing video captioning benchmarks. Most existing video captioning benchmarks, such as MSRVTT~\citep{xu2016msr-vtt}, YouCook2~\citep{zhou2018youcook2} and ActivityNet Caps~\citep{krishna2017activitynet}, also use multi-shot videos as annotation source, but they either annotate a holistic caption for the video (MSRVTT) or ask annotators to decide the boundary of different events. In our study, we observe that video shots naturally create a sequence of related events, motivating us to annotate distinct captions for each shot. Ego4D~\citep{grauman2022ego4d} only annotates dense visual captions but not audio captions for relatively long egocentric videos. Video Storytelling~\citep{li2019video} is a small-scale dataset with annotations of multiple events in a videos and provides a summary of the video by concatenating all captions. 

A recent work  VAST~\citep{chen2024vast} feeds generated video and audio captions into an LLM to generate video summary. However, it processes multi-shot video as a whole and lacks the granularity of the events in different shots. Moreover, VAST directly uses predicted captions without any human verification, leading to potentially noisy and biased summaries towards the captioning models. 
Our dataset stands out from VAST with its accurately annotated visual and narration shot captions. 
Although our video summary is also generated using an LLM, it is further verified by annotators to make sure there is no hallucinated details from the LLM. Our dataset has an average length of $218.3$ words for the video summary, which is much longer than existing benchmarks, and is longer than the combined length of captions in one video in ActivityNet and YouCook2.
%

Furthermore, our Shot2Story-QA introduces unique and complex challenges through its emphasis on shot transitions and multi-event progression, setting it apart from benchmarks like MSRVTT-QA~\cite{xu2017msrvttqa} and ActivityNet-QA~\cite{yu2019activitynetqa}. 
For instance, 
unlike existing benchmarks~\cite{xu2017msrvttqa,yu2019activitynetqa} that assess understanding at a single time point, \eg, "Who do three judges talk to?" (MSRVTT-QA, in Figure~\ref{fig:supp_qa_summary_msrvtt}), or general inquiries like "What is the person in the video doing?" (ActivityNet-QA, in Figure~\ref{fig:supp_qa_summary_anet}), Shot2Story-QA includes "Temporal-related" and "Multi-shot Holistic Understanding" questions. A temporal-related question, such as "What is the man's immediate action after handling the skewer?" shown in Figure~\ref{fig:supp_qa_summary_s2s}, requires models to comprehend the sequence and progression of events, linking consecutive actions meaningfully. Similarly, multi-event progression questions like "How does the setting change from the start to the end of the video?" necessitate understanding multiple concurrent events within their temporal context. This provides a more rigorous and nuanced evaluation framework for temporal understanding.

%% file: sec/experiments.tex
\section{Tasks and Experiments}
\vspace{-2mm}
\label{sec:task}

\begin{table}[]
\caption{Performance of models on video shot captioning using different modalities, following the settings of VAST\citep{chen2024vast}. The models are fine-tuned on Shot2Story video shot captions. V, A and S are abbreviated for vision, audio and subtitle (ASR text) respectively.}
\vspace{-1mm}
\centering
\small
\setlength{\tabcolsep}{6.mm}
\begin{tabular}{@{} l@{} c c l l l l l @{}}
\toprule
Model    &  FT Modality & \multicolumn{1}{c}{B4} & \multicolumn{1}{c}{M} & \multicolumn{1}{c}{R} & \multicolumn{1}{c}{C} \\
\midrule
VAST & V+S & 10.7 & 16.1 & 30.3 & 33.8 \\
VAST & V+A+S & 10.7 & 16.1 & 30.4 & 34.0 \\
MiniGPT4-C & V & 9.2 & 14.7 & 27.9 & 25.1 \\
MiniGPT4-C & V+S & 11.8 & 16.7 & 30.1 & 35.9 \\
VideoChat2-C & V & 8.8 & 16.1 & 27.9 & 23.7 \\
VideoChat2-C & V+S & \textbf{13.8} & \textbf{18.7} & \textbf{32.1} & \textbf{43.9} \\
\bottomrule
\end{tabular}
\vspace{-2mm}
\label{tab:shot_cap}
\end{table}

\begin{table}[]
\caption{Performance of models on video summarization. The models are fine-tuned on Shot2Story video summaries. V and S are abbreviated for vision and subtitle (ASR text) respectively.}
\vspace{-1mm}
\centering
\small
\resizebox{0.85\textwidth}{!}{
\begin{tabular}{@{} l@{} c c l l l l @{}}
\toprule
Model    & FT Modality & \multicolumn{1}{c}{B4} & \multicolumn{1}{c}{M} & \multicolumn{1}{c}{R} & \multicolumn{1}{c}{C} \\
\midrule
Video-ChatGPT w/o ASR ~\citep{maaz2023videochatgpt} & V & 4.8 & 17.3 & 21.3 & 1.5 \\
Video-ChatGPT~\citep{maaz2023videochatgpt} & V+S & 3.6 & 17.8 & 19.7 & 1.0 \\
\midrule
MiniGPT4-SUM-holistic & V+S &  7.8 & 16.9 & 23.4 & 2.8 \\
MiniGPT4-SUM-shot w/o ASR & V &  10.4 & 18.5 & 25.8 & 4.8 \\
MiniGPT4-SUM-shot & V+S &  12.4 & 19.7 & 27.6 & 7.6 \\
VideoChat2-SUM-shot & V+S &  \textbf{12.7} & \textbf{19.8} & \textbf{28.3} & \textbf{9.0} \\
\bottomrule
\end{tabular}
}
\vspace{-2mm}
\label{tab:summary_gen}
\vspace{-2mm}
\end{table}

\subsection{Basic settings}
\vspace{-2mm}
For all the tasks described in this section, we follow the same training/validation/test split. Specifically, the number of videos for training, validation, and test set are 36951, 1982 and 4025, respectively.
We resize the frames to $224\times 224$. We adapt two popular VLMs to accomodate our tasks: MiniGPT-4~\citep{zhu2023minigpt} and VideoChat2~
\citep{li2023mvbench}.
For MiniGPT-4, we employ ViT-G/14~\citep{EVA} and Q-Former~\citep{li2023blip2} as visual encoder, and Vicuna v0-7B~\citep{vicuna2023} as the language model. We load pretrained Q-Former and MLP from MiniGPT-4~\citep{zhu2023minigpt}.
In training, we update only Q-Former and MLP parameters, keeping the ViT and LLM frozen. 
For VideoChat2, we employ UMT-L\citep{li2023unmasked} as backbone and load pretrained Q-Former and MLP from VideoChat2~\citep{li2023mvbench}. During training, we adopt LoRA\citep{hu2021lora} and
AdamW~\citep{loshchilov2017adamw} with a learning rate of 8e-5. We train both models for 10 epochs with a batch size of 128 for single-shot video captioning. 
We finetune our video summarization models on the single-shot captioning models with a batch size of 32. For captioning and summarization, we evaluate the models using BLEU@4~\citep{papineni2002bleu} (B), METEOR~\citep{denkowski2014meteor} (M), ROUGE~\citep{lin2004rouge} (R), and CIDEr~\citep{vedantam2015cider} (C). 

\vspace{-2mm}
\subsection{Single-shot video captioning}
\label{sec:single-shot}
\vspace{-2mm}


This task involves generating descriptions for individual video shots, where the target description is a concatenation of the visual-only and narration caption for a video shot. This task requires a joint understanding of visual and speech information.   Specifically, we adopt a similar structure as we generate pseudo captions for data annotation in Section~\ref{sec:generate_caption}. First, we sample $N_s$ frames from a video shot, encode them using a fixed vision encoder, then feed the encoded features to a Q-Former to produce visual tokens. Further, we combine the visual tokens and an optional ASR text into a unified LLM prompt to facilitate both visual and narration understanding. We adapt the framework of MiniGPT-4~\citep{zhu2023minigpt} and VideoChat2~\citep{li2023mvbench}, with the two models denoted as MiniGPT4-C and VideoChat2-C for brevity. We compare with baseline model VAST~\citep{chen2024vast}, which is able to processes audio, vision, and subtitle inputs simultaneously. 

\begin{figure*}
  \centering
  \includegraphics[width=0.9\linewidth]{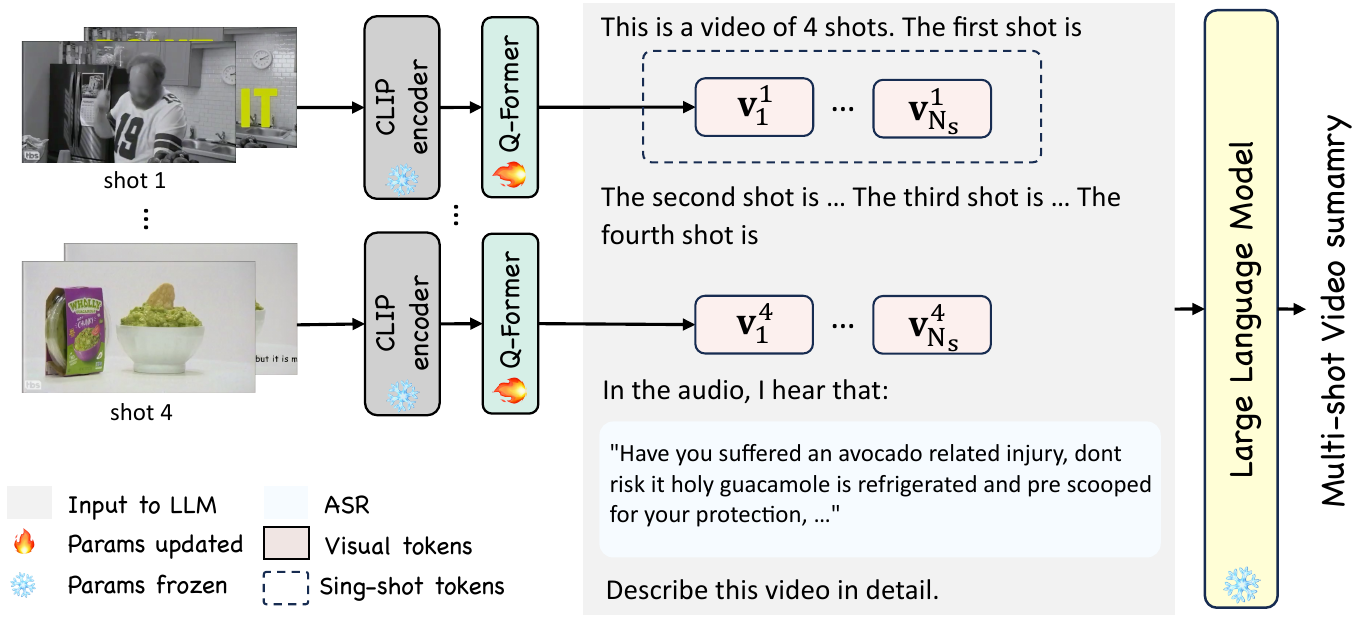}
  \vspace{-2mm}
  \caption{Model structure for multi-shot video summarization model SUM-shot. We arrange visual tokens sequentially for each single shot and in a multi-shot format to encapsulate multi-shot information. Additionally, ASR text is incorporated for audio-visual video summarization.}
  \label{fig:summary_cap}
  \vspace{-3mm}
\end{figure*}

The results are shown in Table~\ref{tab:shot_cap}. Benefiting from extensive pretraining, VAST achieves ~34 on C and 30 on R, comparable to MiniGPT4-C. Incorporating an additional audio modality results in a negligible performance difference, indicating that audio content only contributes marginally given the ASR text as input on our benchmark.
We then assess variants of our models, MiniGPT4-C and VideoChat2-C, with and without the additional ASR text. It shows that including ASR texts significantly enhances performance across all metrics, with a notable boost in R and C, highlighting the relevance of audio content to our video captions.
Furthermore, VideoChat2-C, featuring a superior visual backbone and extensive video pretraining, consistently outperforms MiniGPT4-C. This superiority highlights the importance of advanced visual backbone and video pretraining, confirming our benchmark's robustness. However, despite these advances, the results also indicate room for improvement. Figure~\ref{fig:vis_test} (a) showcases output examples of our model's single-shot video captioning, detailing visual elements and audio content effectively, capturing actions like "gesturing to explain her fear" and secondary elements such as "a red stuffed doll next to her".

\vspace{-2mm}
\subsection{Multi-shot video summarization}
\vspace{-2mm}
\label{sec:multi-summary}
Multi-shot video summarization is a new task that is distinct from existing video description tasks. It requires the model to understand the shot structure of the given video and to provide a coherently paragraph to describe the progression of events in the different shots. Due to the complexity of this task, we adopt GPT-V~\citep{achiam2023gpt} to generate a supplementary training set, with video summaries for another 90K videos, sampled in the same approach as described in Section~\ref{sec:data_preparation}. Please check the annotation prompt and data samples in Appendix~\ref{supp_sec:gptv_prompt}.
First, we finetune an existing video caption model Video-ChatGPT~\citep{maaz2023videochatgpt} by instruction-tuning it on our video summary data, with and without the additional ASR text input.
Then, we experiment with three different architecture designs based on MiniGPT4. The first model MiniGPT4-SUM-holistic uses a similar pipeline as MiniGPT4-C. We uniformly sample 16 frames from the full video clip and prompt the LLM with frame tokens and ASR text. The second model MiniGPT4-SUM-shot w/o ASR, neglecting ASR input, uses a more refined framework by sampling 4 frames in each video shot and prompting the LLM with frame tokens from different shots, as is shown in Figure~\ref{fig:summary_cap}. The third model, MiniGPT4-SUM-shot further incorporates ASR text input as an additional input. Further, we replace the backbone of MiniGPT4-SUM-shot with the more advanced VideoChat2 model, resulting in the VideoChat2-SUM-shot model variant. 
Compared to SUM-shot, SUM-holistic does not have explicit shot information and relies on the LLM to parse the video shots using the provided frame features. 




\begin{figure*}
  \centering
  \includegraphics[width=\linewidth]{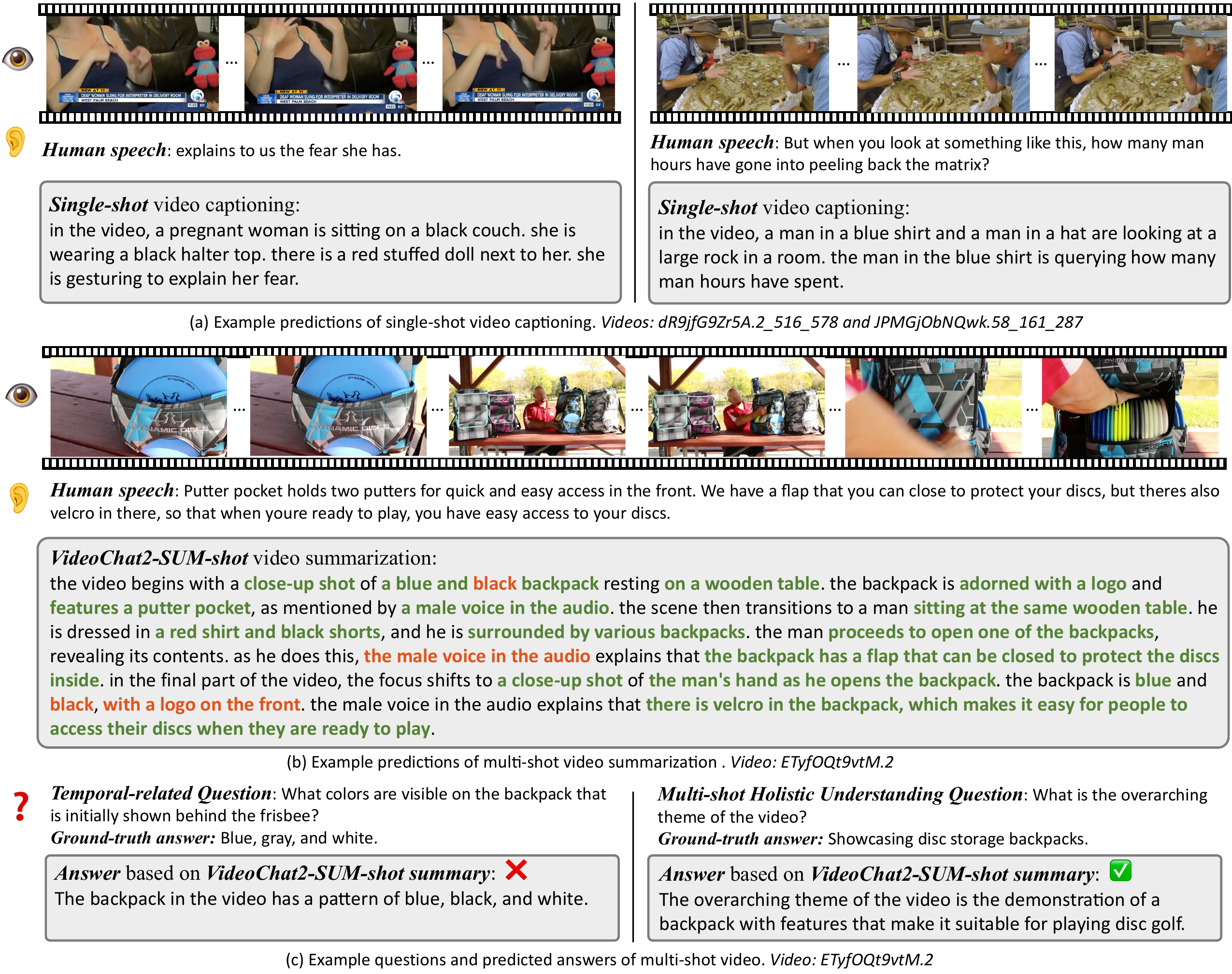}
  \vspace{-5mm}
  \caption{Example predictions of our models. (a) shows single-shot video captioning results of VideoChat2-C, which incorporates audio and visual content correctly (b) shows multi-shot video summarization of VideoChat2-SUM-shot, with accurate descriptions in \textcolor{mygreen}{green} and errors in \textcolor{myred}{red}, illustrating the model's ability to narrate event sequences (c) shows two sample questions of the video in (b). The answers are based on the generated summary of VideoChat2-SUM-shot.}
  \label{fig:vis_test}
  \vspace{-6mm}
\end{figure*}

Table~\ref{tab:summary_gen} shows the results of the models. 
It is shown that MiniGPT4-SUM-holistic is worse than MiniGPT4-SUM-shot, showing the importance of the shot structure in predicting a video summary matching the transition of shots.
MiniGPT4-SUM-shot w/o ASR underperforms compared to MiniGPT4-SUM-shot and outperforms MiniGPT4-SUM-holistic, highlighting the significance of both audio information and shot structure in multi-shot video understanding.
Compared VideoChat2-SUM-shot and MiniGPT4-SUM-shot, the former model achieves the best performance, indicating the benefit of advanced vision backbone and video pretraining. 
Video-ChatGPT obtains much worse performance comparing to our models, potentially due to their weakness in processing multiple scenes and building the correlation between visual frames and ASR texts. It directly encodes the whole video into a sequence of tokens, potentially losing significant frame details and essential correlation between ASR and visual frames, while ours directly feed frames tokens into the LLM without compressing them.


Figure~\ref{fig:vis_test} (b) showcases predictive capabilities of our VideoChat2-SUM-shot model. The model adeptly narrates event sequences with appropriate emphasis. For instance, it details the backpack's colour and  location, and rationalizes the item in the beginning shot, \ie, ``putter pocket'', aligning with the ASR with ``by a male voice in the audio''. However, some predictions that marked in red are erroneous, such as the incorrect ``black'' colour and the non-existent ``with a logo on the front'' in the ending shot. 
These inaccuracies likely stem from the LLM's tendency to ``hallucinate'' plausible yet non-factual details. Despite these errors, the model demonstrates proficiency in generating consistent and nuanced summaries, highlighting our model's potential and the challenges our dataset presents.



\subsection{Video question-answering with video summary}
\label{sec:video_qa_text}
Generated video summaries are supposed to be grounded and detailed, covering rich elements like event progression, holistic topics and audio elements,
making them suitable for other vision tasks such as video question-answering.
Existing work~\citep{guo2023imagestextualprompt,zhang2023simple} uses image or video frame captions as input to an LLM to generate question responses.  However, little work has been done for the capacity of video summaries.
We directly apply our video summarization model on video QA benchmarks, \ie MSRVTT-QA~\citep{xu2017msrvttqa}, ActivityNet-QA~\citep{yu2019activitynetqa} and our Shot2Story-QA.  

Specifically, we first split the testing videos into video shots, and then feed the videos into our SUM-shot models. The generated summaries and the associated questions are then fed into a Vicuna model to derive the answers with the prompt shown in Appendix~\ref{supp_sec:zr_qa_vicuna_prompt}. Note there is no adaptation or finetuning conducted for the Vicuna model.
Since the original answers in the QA benchmarks are very short and the responses generated by LLM tend to be long sentences, we leverage the \texttt{gpt-3.5-turbo} model to generate a binary decision on whether the answer is correct, following Video-ChatGPT~\citep{maaz2023videochatgpt}. 

\begin{figure*}
  \centering
  \includegraphics[width=\linewidth]{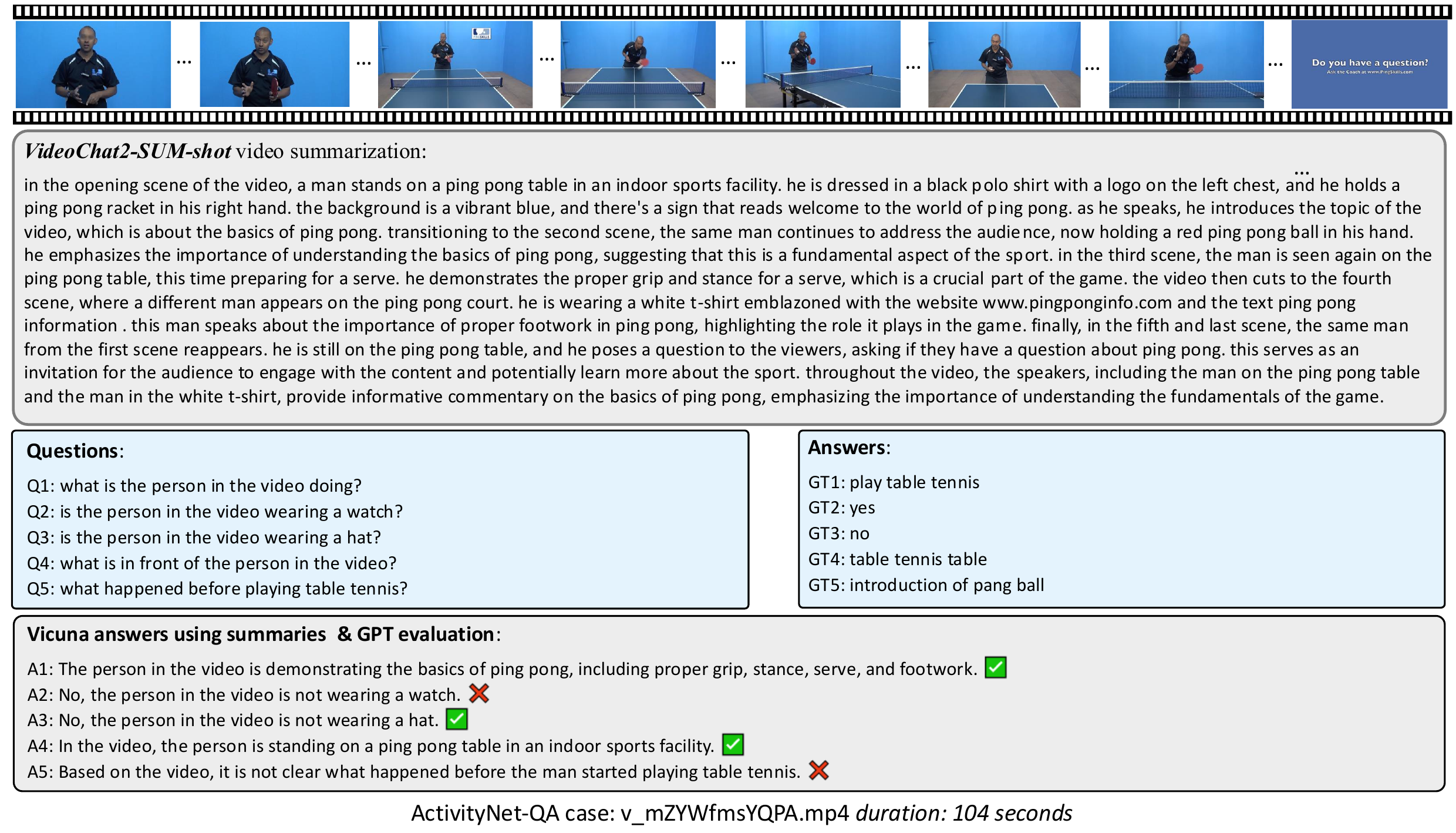}
  \caption{Example predictions of our model on zero-shot question answering an ActivityNet-QA video. More questions and explanations can be found in Appendix~\ref{supp_sec:qa_summary_examples}.}
  \label{fig:vis_zr_qa_test}
  \vspace{-4mm}
\end{figure*}

\textbf{Zero-shot video question-answering.} 
As shown in Table~\ref{tab:zr_qa}, our results with VideoChat2-SUM-shot surpass 5 out of 6 existing video-VLMs on MSRVTT-QA and 4 out of 6 existing models on ActivityNet-QA. 
Furthermore, our results are comparable to the SOTA performance on MSRVTT-QA with Video-LLaVA~\citep{lin2023video}. 
Note that these models require extensive instruction-tuning data to learn to directly generate answers from visual features and the text prompt, whereas our model bypasses instruction tuning by distilling the video information into a summary. 
Our model also follows the zero-shot QA setting since the model only uses \dataset as training data. Note that MSRVTT contains a large portion of videos with out-of-domain topics such as TV shows (e.g., Figure~\ref{fig:supp_qa_summary_msrvtt}) and food, while ActivityNet has much longer videos than our training videos (e.g., Figure~\ref{fig:vis_zr_qa_test}. This validates the robustness and transferability of our model across different topics and longer videos. This surprisingly good result indicates that a comprehensive and detailed video summary is a high-quality abstraction of the video, facilitating a wide range of tasks including video QA and video-based conversation. 
Moreover, our model achieves strong results on ActivityNet-QA, which predominantly consists of single-shot long videos, validating that models trained with multi-shot videos can effectively generalize to single-shot videos.

\textbf{Multi-shot video question-answering.} As shown in Table~\ref{tab:id_qa}, we benchmark existing and our proposed video summary models on Shot2Story-QA. Specifically, four popular video-VLMs are compared, \ie, Video-ChatGPT~\citep{maaz2023videochatgpt}, LLaMA-VID~\citep{li2023llama}, VideoChat2~\citep{li2023mvbench} and Video-LLaVA~\citep{lin2023video}. The predicted summaries of MiniGPT4-SUM-shot and VideoChat2-SUM-shot are used to tackle the QA task with the same configuration as zero-shot VQA.
Accuracies on temporal-related, multi-shot holistic understanding and audio-related are reported, with the ``overall'' metric showing the average score from these three sub-tasks. The current video-VLMs present unsatisfying results, potentially due to two factors: (1) The current models does not have audio or ASR as input, lacking capacity with audio-related understanding. (2) Current models do not have training data with detailed descriptions based on multi-shot videos, weakening their performance on holistic understanding and temporal modeling. 
For our models, VideoChat2-SUM-shot achieves an overall score of 40.5, surpassing the compared models and MiniGPT4-SUM-shot on all three subtasks. This performance underscores the benefits of video pretraining and the advanced visual backbone of VideoChat2. These baseline results highlight the complexities and demanding nature of our Shot2Story-QA task. We show some example predictions in Figure~\ref{fig:vis_test}(c). Please refer to Appendix~\ref{supp_sec:qa_summary_examples} for more qualitative results.

\label{sec:zr_video_qa}

\begin{table}
\caption{Performance on video question answering on MSRVTT-QA and ActivityNet-QA. IT means whether the model uses video-text instruction tuning data. All methods follow the zero-shot manner. }
\vspace{-2mm}
\centering
\small
\setlength{\tabcolsep}{3.5mm}
\begin{tabular}{@{} c | l | c | c | c | c @{}}
\toprule
\multirow{2}{*}{Model} & \multirow{2}{*}{Training Data} & \multirow{2}{*}{IT} & \multirow{1}{*}{QA}  & MSRVTT & ActivityNet  \\ 
                       &      &    & Input & QA & QA \\
\midrule
VideoChat~\citep{li2023videochat} & Cap.+QA & \cmark & V+T &45.0 & 26.5 \\
Video-ChatGPT~\citep{maaz2023videochatgpt} & Cap.+QA &  \cmark & V+T & 49.3 & 35.2 \\
MovieChat~\citep{song2023moviechat} & Cap.+QA & \cmark &  V+T & 52.7 & 45.7 \\
LLaMA-VID~\citep{li2023llama} & Cap.+QA & \cmark &  V+T & 57.7 & 47.4 \\
VideoChat2~\citep{li2023mvbench} & Cap.+QA & \cmark &  V+T & 54.1 & \textbf{49.1} \\
Video-LLaVA~\citep{lin2023video} & Cap.+QA & \cmark &  V+T & \textbf{59.2} & 45.3 \\
\midrule

MiniGPT4-SUM-shot & Summary & \xmark & T & 57.7 & 45.6 \\
VideoChat2-SUM-shot & Summary & \xmark & T & 58.5 & 47.1 \\
\bottomrule
\end{tabular}
\label{tab:zr_qa}
\vspace{-2mm}
\end{table}

\begin{table}
\caption{\textcolor{black}{Benchmark on Shot2Story-QA.} 
IT means usage of video-text instruction tuning data. 
Summary, Cap. and QA denote video summary, captions and question-answering pairs.}
\vspace{-2mm}
\centering
\small
\setlength{\tabcolsep}{1.3mm}
\begin{tabular}{@{} c | l | c | c | c | c | c | c @{}}
\toprule
\multirow{2}{*}{Model} & \multirow{2}{*}{Training data} & \multirow{2}{*}{IT} & \multirow{1}{*}{QA}  & Temporal& \multirow{1}{*}{Holistic}  & Audio & \multirow{2}{*}{Overall}\\ 
                       &   &   & Input & related & understanding & related &  \\ 
\midrule
LLaMA-VID~\citep{li2023llama} & Cap.+QA & \cmark &  V+T & 7.9 & 9.7 & 11.4 & 9.7  \\
Video-ChatGPT~\citep{maaz2023videochatgpt} & Cap.+QA &  \cmark & V+T & 13.1 & 15.5 & 14.3 & 14.2 \\
VideoChat2~\citep{li2023mvbench} & Cap.+QA & \cmark &  V+T & 15.1 & 15.4 & 13 & 14.5 \\
Video-LLaVA~\citep{lin2023video} & Cap.+QA & \cmark &  V+T & 16.4 & 14.8 & 11.7 & 14.3 \\
\midrule
MiniGPT4-SUM-shot & Summary & \xmark &  T & 28.9 & 31.9 & 36.7 & 32.5 \\
VideoChat2-SUM-shot & Summary & \xmark &  T & 36.1 & 41.5 & 43.8 & 40.5 \\

\bottomrule
\end{tabular}
\label{tab:id_qa}
\vspace{-4mm}
\end{table}

%% file: sec/conclusion.tex
\vspace{-2mm}
\section{Conclusion}
\vspace{-2mm}

In this work, we present Shot2Story, a large-scale benchmark for comprehensive multi-shot video understanding. We provide detailed shot-level captions for both visual signals and human narrations. Furthermore, we provide comprehensive video summaries based on shot-level captions and design a challenging video question-answering benchmark for multi-shot video understanding. 
With the rich and diverse descriptions, our benchmark serves as a playground for future multi-modal video understanding models, ready to be extended for a range of other video understanding tasks, such as visual grounding and video-based conversation.

%% file: sec/X_suppl.tex
\clearpage
\setcounter{page}{1}






\section{Annotation Process}
\label{supp_sec:anno_process}
In this section, the annotation process for \dataset is detailed, including single-shot caption annotation in Sec.~\ref{supp_sec:human_anno_shot}, GPT-4 summary generation prompts in Sec.~\ref{supp_sec:gpt4_prompt}, human correction of summaries in Sec.~\ref{supp_sec:human_anno_sum}, and annotation of question-answering pairs in Sec.~\ref{supp_sec:anno_qa}.

\begin{figure*}[h]
  \centering
  \includegraphics[width=0.95\linewidth]{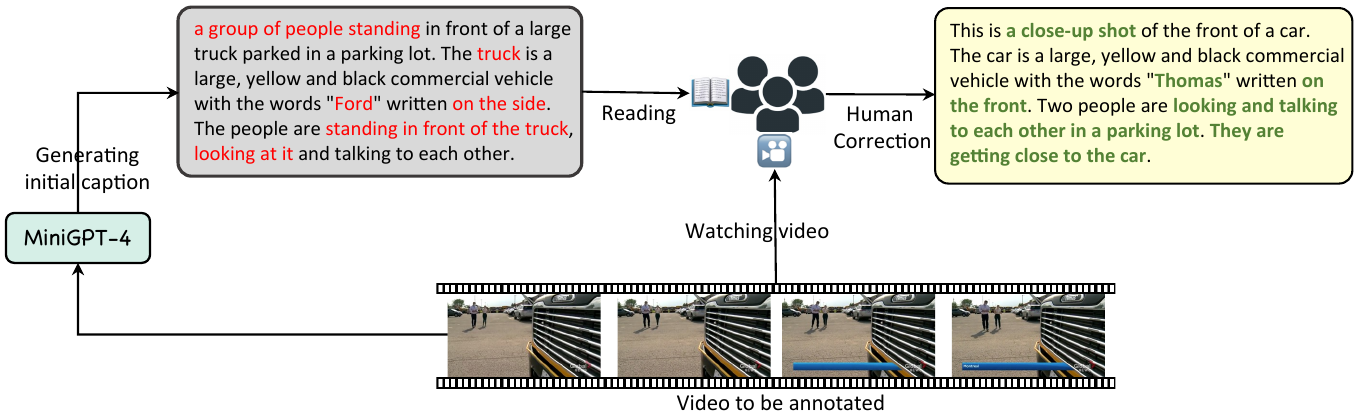}
  \caption{Human annotation process of sing-shot video captions. Texts in bold green represent correct content, while those in red indicate errors. Please find more explanations in Sec.~\ref{supp_sec:human_anno_shot}.
  }
  \label{fig:supp_anno_process}
  \vspace{-2mm}
\end{figure*}

\subsection{Human annotation of single-shot captions}
\label{supp_sec:human_anno_shot}
Our single-shot video caption annotation process, described in Sec.~\ref{sec:generate_caption}, is a two-phase approach designed for high-quality and style-consistent captions. This procedure also accelerates the annotation process by $\sim3$ times. As depicted in Figure~\ref{fig:supp_anno_process}, the process begins with MiniGPT-4 generating initial captions from structured prompts. While these captions often correctly identify subjects such as ``parking lot'' and ``vehicle'', they sometimes inaccurately describe actions or locations. Annotators then watch the video shot and revise these captions. For instance, errors like ``standing in front of'', depicted in red, are corrected to ``getting close to the car'', shown in green. Additionally, annotators enrich captions with key details, such as ``a close-up shot of the front of a car.''
Our single-shot narration caption annotation process follows a similar approach. Differently, we offer ASR text and videos to the annotators and ask them to write down the visually related content and describe the source of the speech. The process during model deployment has been detailed in Sec.~\ref{sec:single-shot}.

\begin{figure*}
  \centering
  \includegraphics[width=0.75\linewidth]{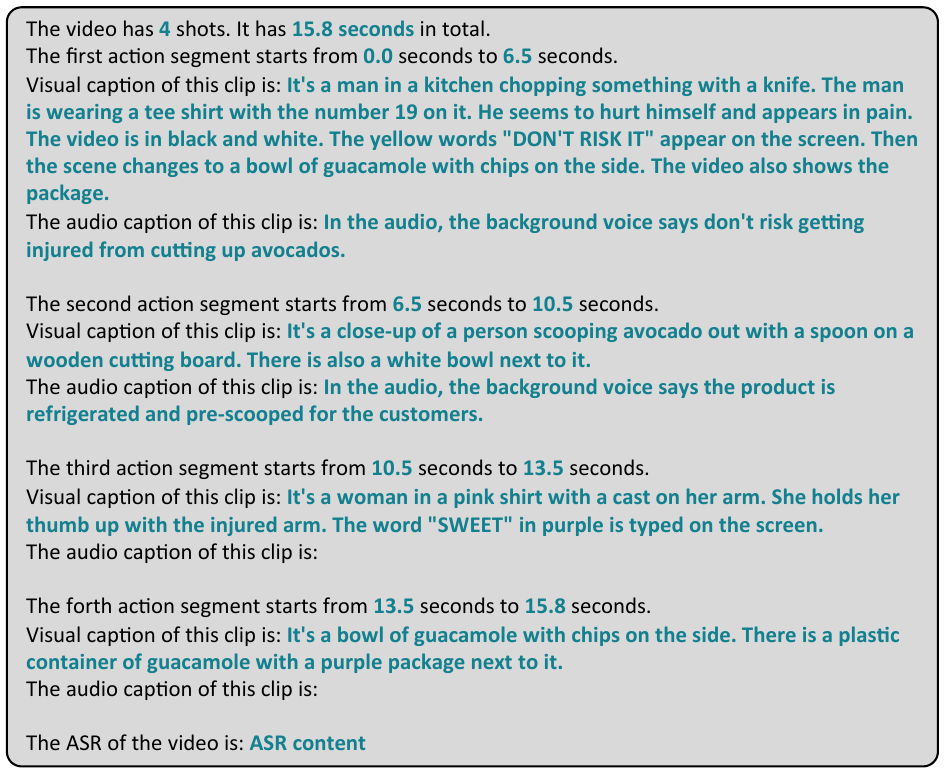}
  \caption{Example of textual content for video in Figure~\ref{fig:video_demo}. Texts in color are specific for input video and are replaced during our generation.
  }
  \label{fig:supp_video_content}
  \vspace{-2mm}
\end{figure*}

\subsection{GPT-4 summarization prompt}
\label{supp_sec:gpt4_prompt}

\begin{figure*}[ht]
  \centering
  \includegraphics[width=0.8\linewidth]{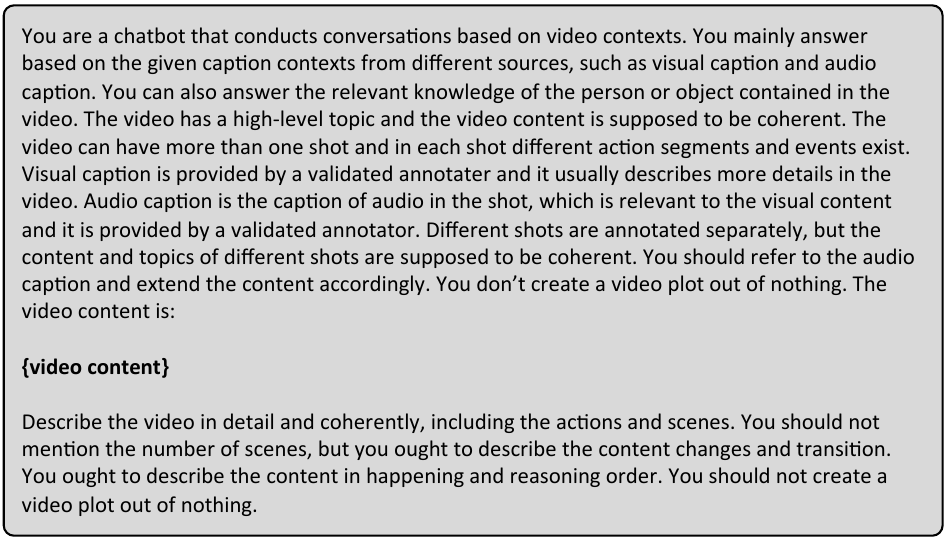}
  \caption{Prompt template for GPT-4 summarization. 
  }
  \label{fig:supp_gpt4_template}
  \vspace{-2mm}
\end{figure*}

We utilize GPT-4 to summarize video clips, leveraging our detailed video-shot captions and ASR text. The summarization follows a prompt structure adapted from \citep{li2023videochat}, which defines video captions and audio captions for each shot, as depicted in Figure~\ref{fig:supp_gpt4_template}. For each video, we organize shot durations, video captions, narration captions, and ASR into a text format (see Figure~\ref{fig:supp_video_content}). This arranged content is then fed into GPT-4 for generating the video summary.

\subsection{Human correction of video summaries}
\label{supp_sec:human_anno_sum}
\begin{figure*}[t!]
  \centering
  \includegraphics[width=0.98\linewidth]{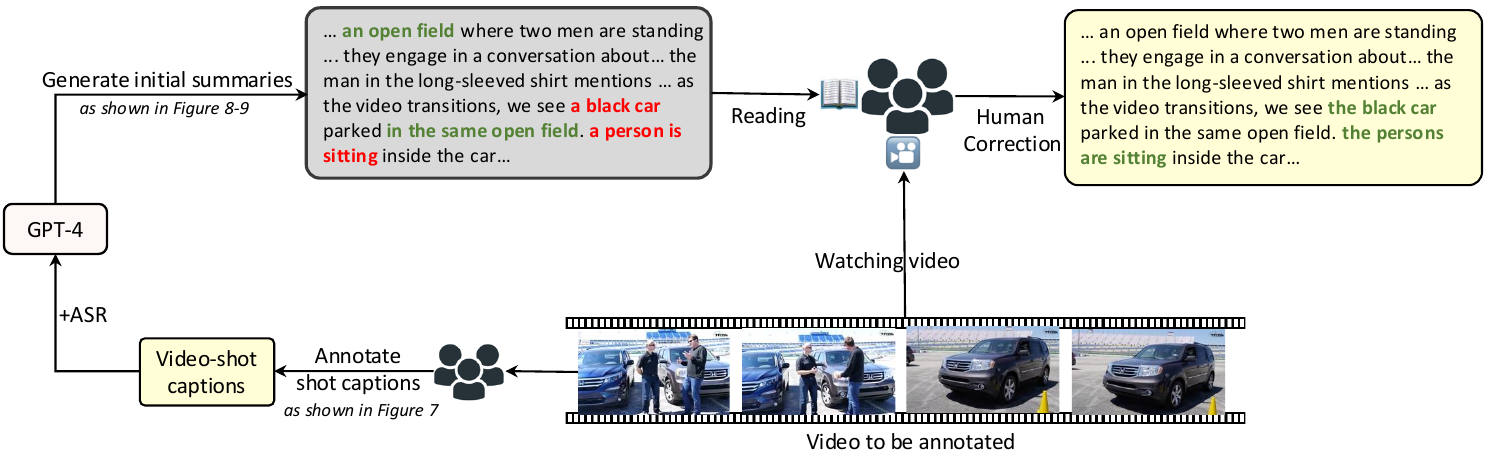}
  \caption{Human correction process of video summaries. Overlapped text is omitted for clarity. Texts in bold green represent correct content, while those in red indicate errors. Please find more explanations in Sec.~\ref{supp_sec:human_anno_sum}.
  }
  \label{fig:supp_refer_correct}
  \vspace{-2mm}
\end{figure*}

Our detailed shot captions enable GPT-4 to effectively identify and link subjects across shots, without requiring extra re-identification modules. However, according to human evaluation, about 30\% of video summaries struggle to connect objects and scenes across shots. Our annotators review these summaries alongside the video clips to correct such errors. 
Figure~\ref{fig:supp_refer_correct} illustrates this process. While GPT-4 accurately references the same location, such as ``the open field'' and ``the same open field'', it sometimes fails to maintain continuity with elements like ``the black car'' across scene transitions. Annotators must watch the video and assess the initial summary to make necessary corrections for the final summary. This method ensures the production of high-quality video summaries with efficiency.

\subsection{Annotation of video question-answering pairs}
\label{supp_sec:anno_qa}
\begin{figure*}[ht]
  \centering
  \includegraphics[width=\linewidth]{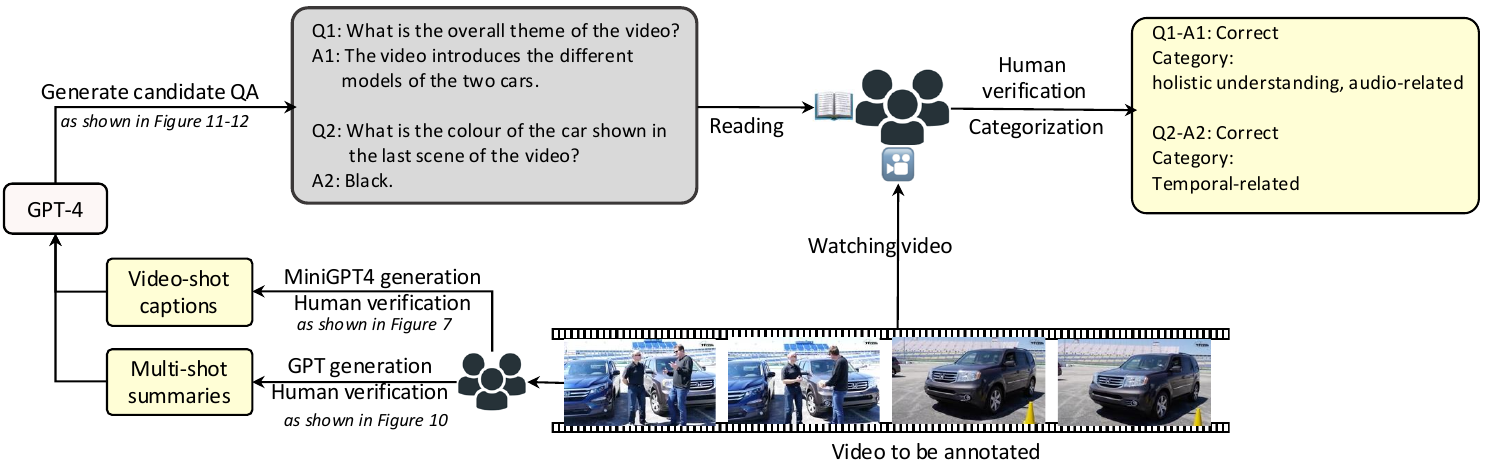}
  \caption{Human annotation process of QA pairs. Please find more explanation in Sec.~\ref{supp_sec:anno_qa}. 
  }
  \label{fig:supp_gpt4_qa_anno}
  \vspace{-2mm}
\end{figure*}

As shown in Figure~\ref{fig:supp_gpt4_qa_anno}, our annotation of question-answering (QA) pairs utilizes a hybrid manual-automatic approach to ensure both diversity and quality. Initially, we employ GPT-4 to generate candidate question-answer pairs based on our specific instructions, focusing on creating temporal-related, holistic understanding, and audio-related questions. Subsequently, annotators review these pairs while watching the corresponding videos to eliminate simple or incorrect entries. Subsequently, the annotators are asked to categorize the QA pairs into predefined categories, \ie, temporal-related, holistic understanding and audio-related, and discard the data not under these categories. 

\begin{figure*}[t!]
  \centering
  \includegraphics[width=\linewidth]{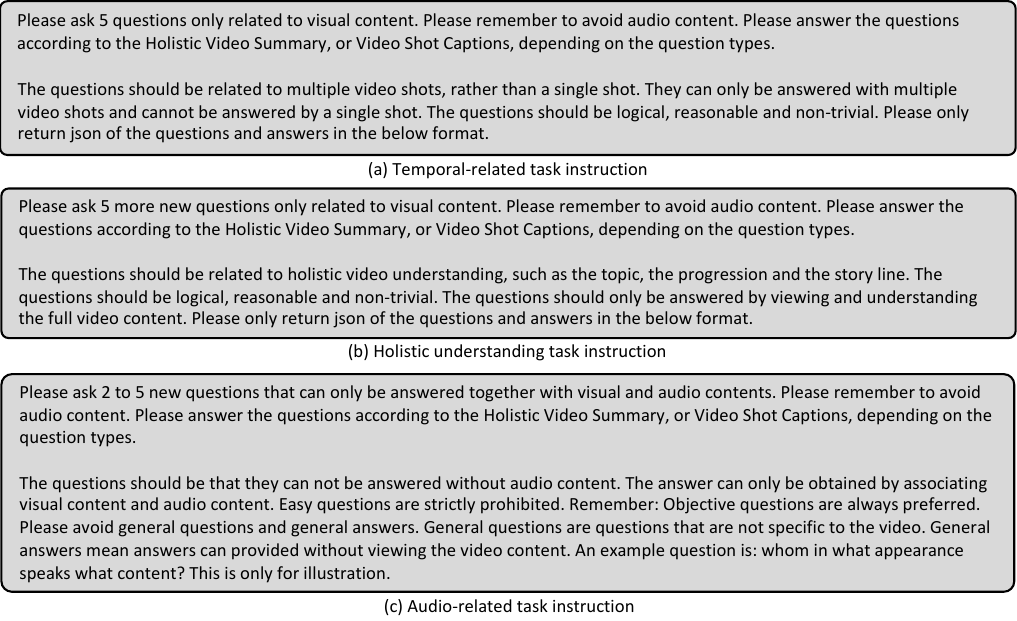}
  \caption{Task instruction for candidate question-answering pairs generation. 
  }
  \label{fig:supp_gpt4_qa_task_inst}
  \vspace{-2mm}
\end{figure*}

\begin{figure*}[ht]
  \centering
  \includegraphics[width=\linewidth]{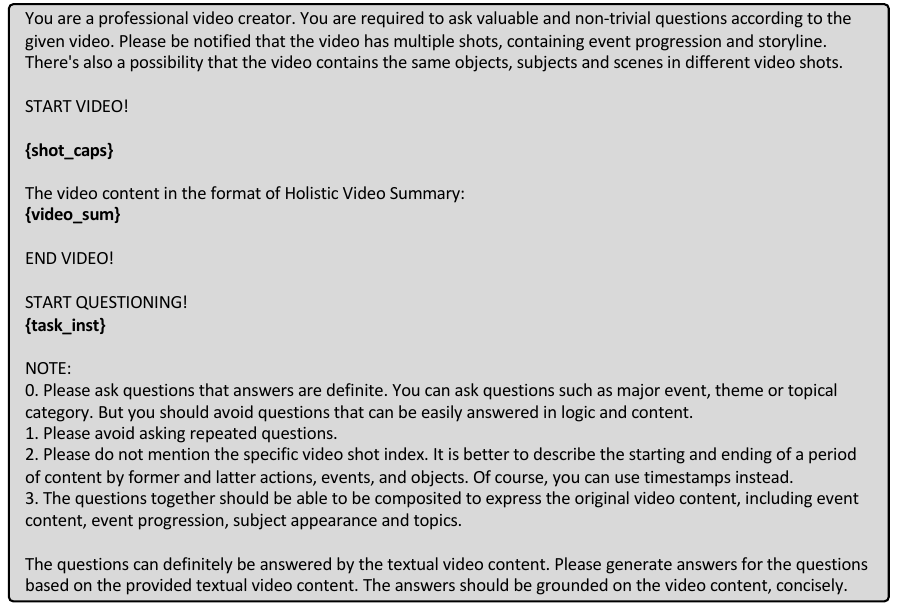}
  \caption{Prompt template for candidate question-answering pairs generation. 
  }
  \label{fig:supp_gpt4_qa_generate}
  \vspace{-2mm}
\end{figure*}

The prompts used for generating candidate QA pairs are detailed in Figure~\ref{fig:supp_gpt4_qa_task_inst} and Figure~\ref{fig:supp_gpt4_qa_generate}. For different task-specific questions, we adopt different task instructions as shown in Figure~\ref{fig:supp_gpt4_qa_task_inst}. During generation, we adopt the template shown in Figure~\ref{fig:supp_gpt4_qa_generate}. The boldfaced texts, such as ``\textbf{\{shot\_caps\}}'', ``\textbf{\{video\_sum\}}'', ``\textbf{\{task\_inst\}}'', are replaced with shot captions organized as in Figure~\ref{fig:supp_video_content}, video summary and task-specific instructions from Figure~\ref{fig:supp_gpt4_qa_task_inst}. 

\section{Prompts Used in Our Models}
\label{supp_sec:our_prompt_all}
In this section, we elaborate on the prompts used for training and testing our models. We detail the prompts for single-shot video captioning and narration captioning in Sec.~\ref{supp_sec:single_shot_cap_prompt}. For video summarization models like SUM-shot, SUM-holistic, and SUM-text, the prompts are thoroughly explained in Sec.~\ref{supp_sec:single_video_sum_prompt}.

\subsection{Prompt for single-shot video captioning}
\label{supp_sec:single_shot_cap_prompt}
During the training of our single-shot captioning models, we select a random text prompt for each video shot, with different model variants utilizing distinct prompts. The prompts for the single-shot video captioning model that incorporates both visual signals and ASR are depicted in Figure~\ref{fig:supp_shot_caption_prompt}. In the figure, boldfaced text, such as ``\textbf{\{asr\}}'', is replaced with specific video information. The arrangement of visual tokens and text prompts, as presented in Figure~\ref{fig:summary_cap}, is not included here for brevity.



\subsection{Prompt for video summarization}
\label{supp_sec:single_video_sum_prompt}
\begin{figure*}[t!]
  \centering
  \includegraphics[width=0.8\linewidth]{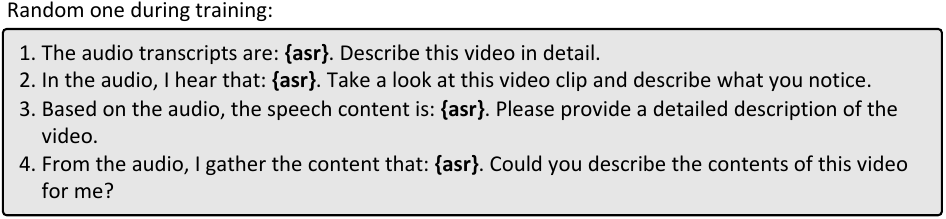}
  \caption{Prompt during training for single-shot video captioning.
  }
  \label{fig:supp_shot_caption_prompt}
  \vspace{-2mm}
\end{figure*}

In Sec.~\ref{sec:multi-summary}, we explore different model variants for video summarization, namely MiniGPT4-SUM-shot, MiniGPT4-SUM-holistic, and VideoChat2-SUM-shot. For training these models, we use the same text prompt as in single-shot video captioning, shown in Figure~\ref{fig:supp_shot_caption_prompt}. The key distinction between MiniGPT4-SUM-shot and MiniGPT4-SUM-holistic lies in the arrangement of visual tokens: MiniGPT4-SUM-shot incorporates shot-specific information such as shot number or index along with visual tokens from each shot, whereas MiniGPT4-SUM-holistic uniformly samples frames across the video. 
For VideoChat2-SUM-shot, the input prompt is the same to MiniGPT4-SUM-shot. 


\subsection{Prompt for in-domain video question-answering}
\label{supp_sec:prompt_qa}
In our paper, we propose a unique question-answering procedure in which we generate video summaries and prompt an LLM to answer the corresponding question. The text prompt used for LLM is shown in Figure~\ref{fig:supp_qa_prompt}. 

\subsection{Prompts used for zero-shot QA}
\label{supp_sec:zr_qa_prompt}
In this section, we detail the prompts employed for the zero-shot video QA task, which is discussed in Sec.~\ref{sec:video_qa_text}. Note that we use the same prompt with Vicuna for text summary-based video question-answering, both zero-shot video QA and in-domain video QA. 

\vspace{-2mm}
\subsubsection{LLM QA prompt}
\label{supp_sec:zr_qa_vicuna_prompt}
\begin{figure*}
  \centering
  \includegraphics[width=0.8\linewidth]{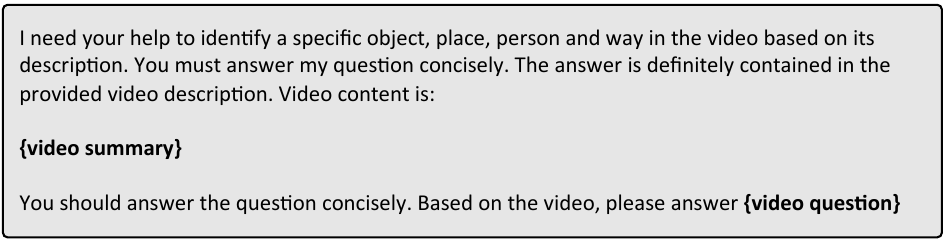}
  \caption{Prompt used for video question answering with summaries.
  }
  \label{fig:supp_qa_prompt}
  \vspace{-2mm}
\end{figure*}

The text prompt is shown in Figure~\ref{fig:supp_qa_prompt}, the same to in-domain video QA. It requires video summaries, which are generated from models trained with our Shot2Story data, which fullfils the definition of zero-shot tasks.
To better align with the ground truth answers, we prompt the LLM to generate concise answers solely based on the provided video content.

\vspace{-2mm}
\subsubsection{Evaluation prompt}
\label{supp_sec:zr_qa_eval_prompt}
In our paper, we follow the same evaluation procedure as outlined in \citep{maaz2023videochatgpt}, using ChatGPT-3.5 to assess the alignment of the generated answers with the given ground truth. 

\subsection{GPTV generation prompt}
\label{supp_sec:gptv_prompt}
\begin{figure*}[t]
  \centering
  \includegraphics[width=\linewidth]{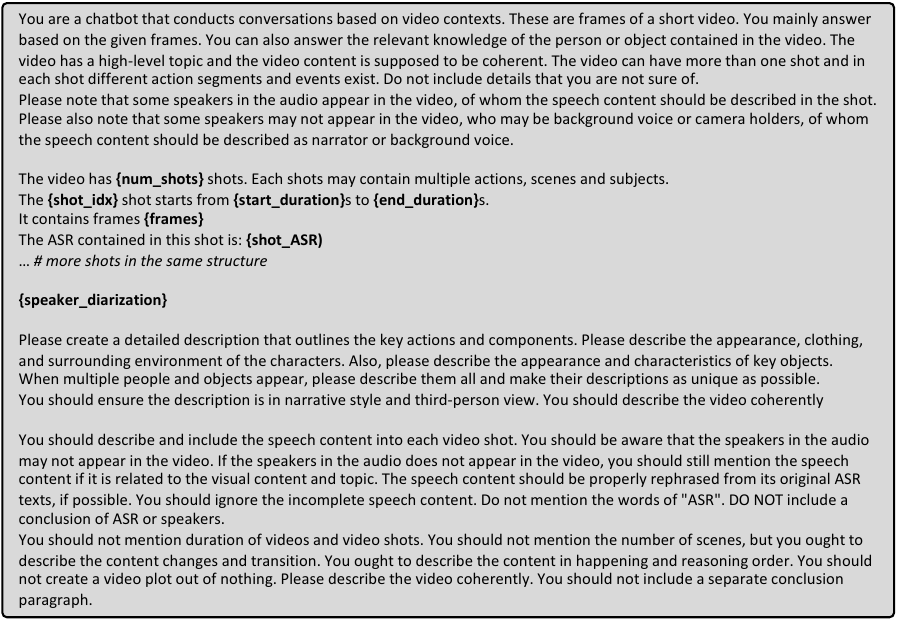}
  \caption{Prompt for video summary generation using GPTV.
  }
  \label{fig:supp_gptv_prompt}
  \vspace{-2mm}
\end{figure*}

\begin{figure*}[t]
  \centering
  \includegraphics[width=\linewidth]{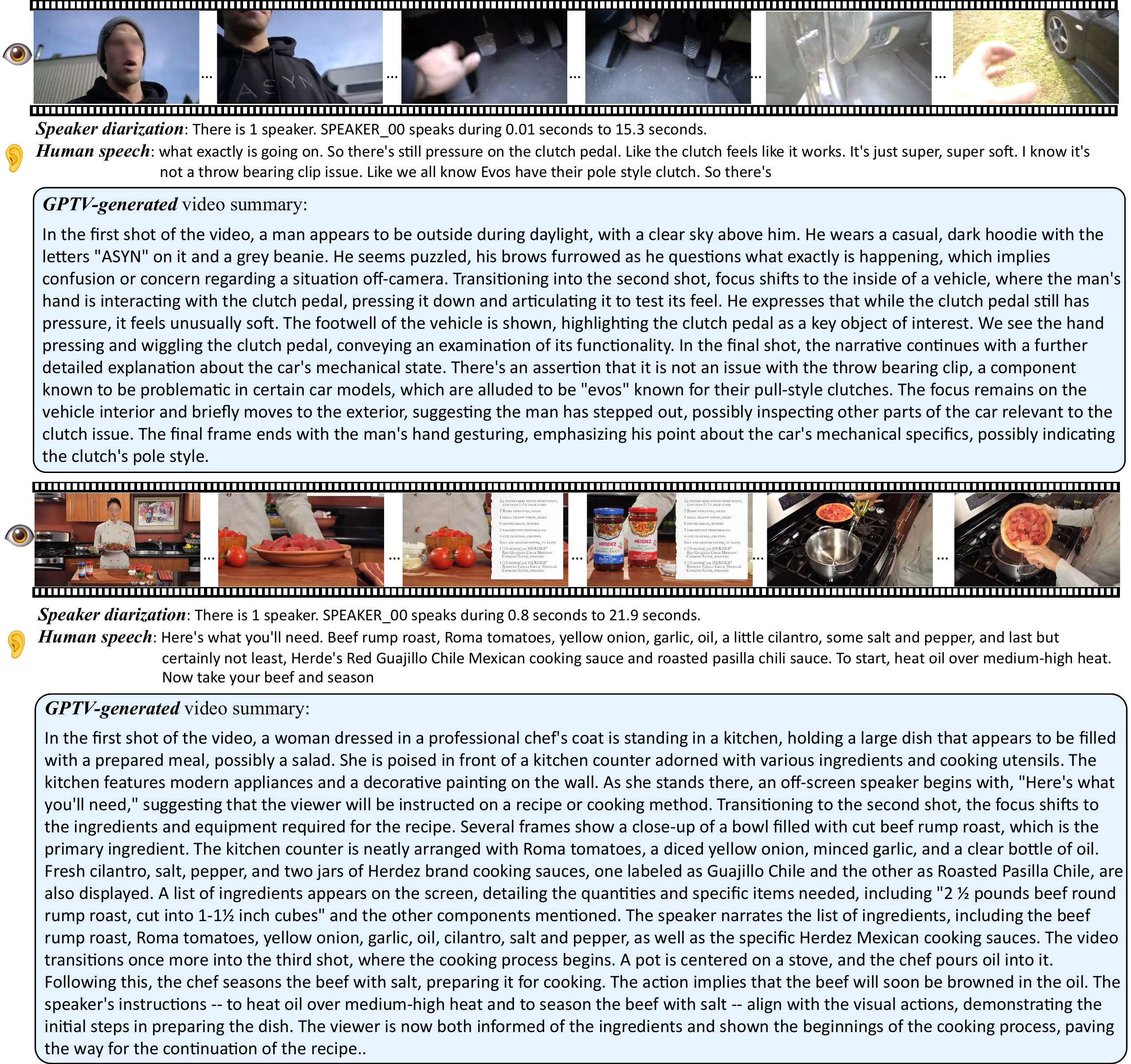}
  \caption{Samples of video summaries generated by GPTV. In both cases, GPTV successfully correlates the audio speaker and the person in the video, \ie ``He expresses that'' and ``an off-screen speaker begins''. Moreover, GPTV-generated summaries capture the overall action and event flow with the help of shot structure, and the essential details in the video.
  }
  \label{fig:supp_gptv_samples}
  \vspace{-2mm}
\end{figure*}

The prompt used for GPTV generation for another 90K videos is shown in Figure~\ref{fig:supp_gptv_prompt}. We follow a similar prompt structure as Figure~\ref{fig:supp_video_content} to organize the different shots. In addition to the shot structure, we additionally embed speaker information and ASR texts for visual-audio correlation in generated summaries. Using Whisper-X, we extract the speaker diarization and ASR texts, which are organized in the form of ``\textit{somebody} speaks \textit{something} during \textit{when} to \textit{when}''. The speaker diarization descriptions are then appended to the tail of video content in the prompt, in order to have speaker identification information in the generated summaries.
We show two samples in Figure~\ref{fig:supp_gptv_samples}, which shows the successful speaker assignment to the visual elements and adequate details. Note this subset of data is not verified by human annotators. We only used it in the video summarization experiments in Section 3.3 of the main paper.

\section{Additional Experiment Results}
\label{supp_sec:examples}
\subsection{Comparison of different models}
\label{supp_sec:comparison_examples}
\begin{figure*}
  \centering
  \includegraphics[width=0.98\linewidth]{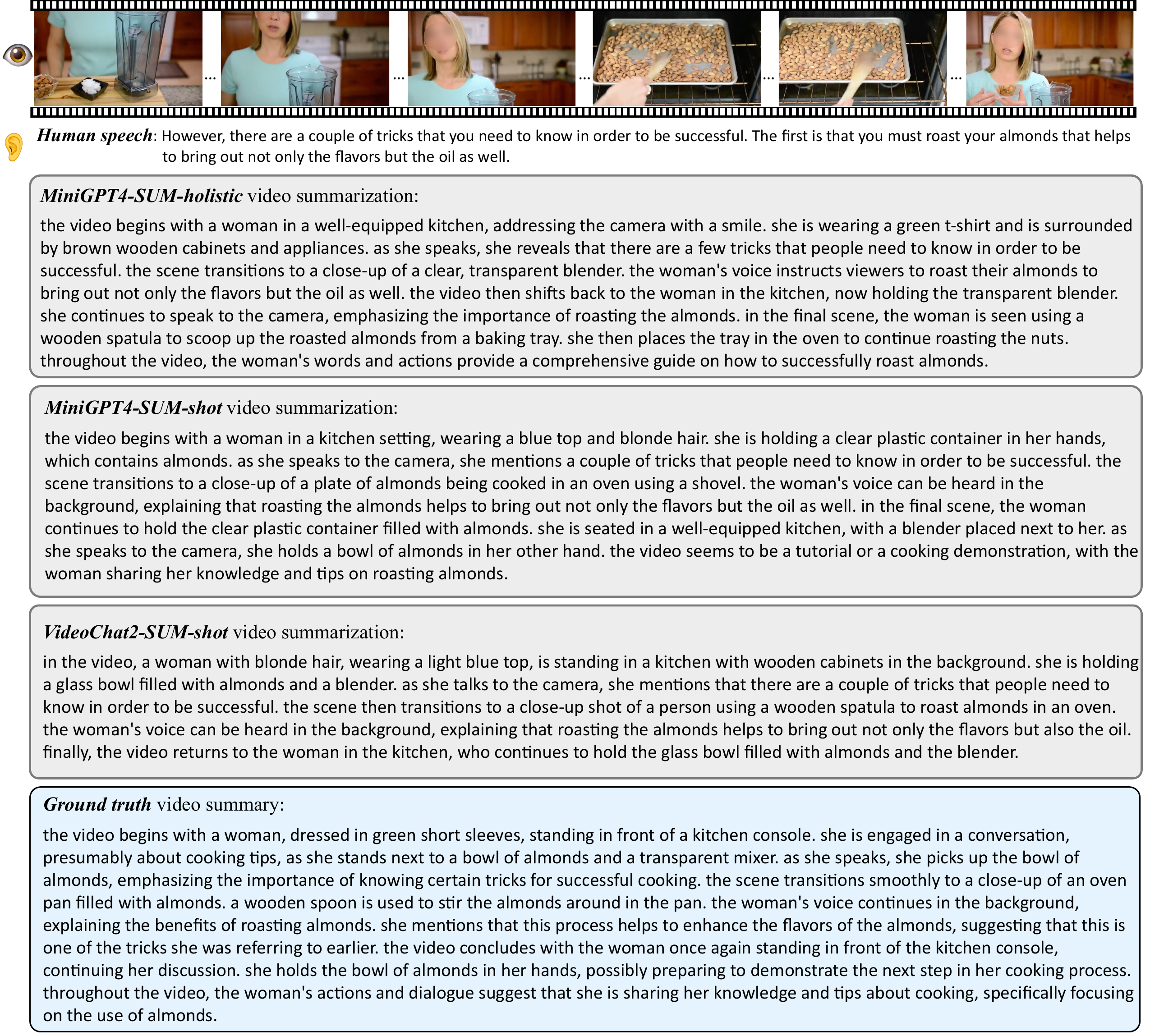}
  \caption{\textcolor{black}{Example for video summarization using MiniGPT4-SUM-holistic, MiniGPT4-SUM-shot and VideoChat2-SUM-shot. 
  All three models effectively grasp the video's main topic and content. Please find explanations in Sec.~\ref{supp_sec:comparison_examples}.}}
  \label{fig:supp_sum_summary_sample1}
  \vspace{-2mm}
\end{figure*}

In this subsection, we present an example of video summarization from our \dataset testing split in Figure~\ref{fig:supp_sum_summary_sample1}, using MiniGPT4-SUM-holistic, MiniGPT4-SUM-shot, and VideoChat2-SUM-shot. Both MiniGPT4-SUM-shot and VideoChat2-SUM-shot, with their access to shot information including shot count and visual tokens in shots, successfully capture the video's storyline and transitions. For instance, they accurately depict the sequence involving a woman in the kitchen, almonds in an oven, and the woman speaking to the camera. VideoChat2-SUM-shot, with its advanced vision backbone and video pretraining, captures more nuanced action details, like ``using a wooden spatula to roast almonds in an oven''.
MiniGPT4-SUM-holistic, while effectively identifying major content and events, falls short in accurate scene sequencing due to its lack of shot-structured visual tokens. This leads to errors in narrative order, such as reversing the scenes of ``shifts back to the woman in the kitchen'' and ``in the final scene, the woman is seen using a wooden spatula''.

\subsection{QA summary}
\label{supp_sec:qa_summary_examples}

In this subsection, we present the results of zero-shot video question-answering using Vicuna v0-13B, based on textual summaries of video samples from MSRVTT-QA~\citep{xu2017msrvttqa} and ActivityNet-QA~\citep{yu2019activitynetqa}. Despite the limitations of our summarization model, which scores 8.6 in CIDEr on the \dataset test split (see Table~\ref{tab:summary_gen}), and the inherent challenges of the videos due to out-of-domain topics or extended durations, the summaries generated from our trained SUM-shot model largely succeed in capturing the key elements of the videos and providing relevant information.

\begin{figure*}
  \centering
  \includegraphics[width=0.98\linewidth]{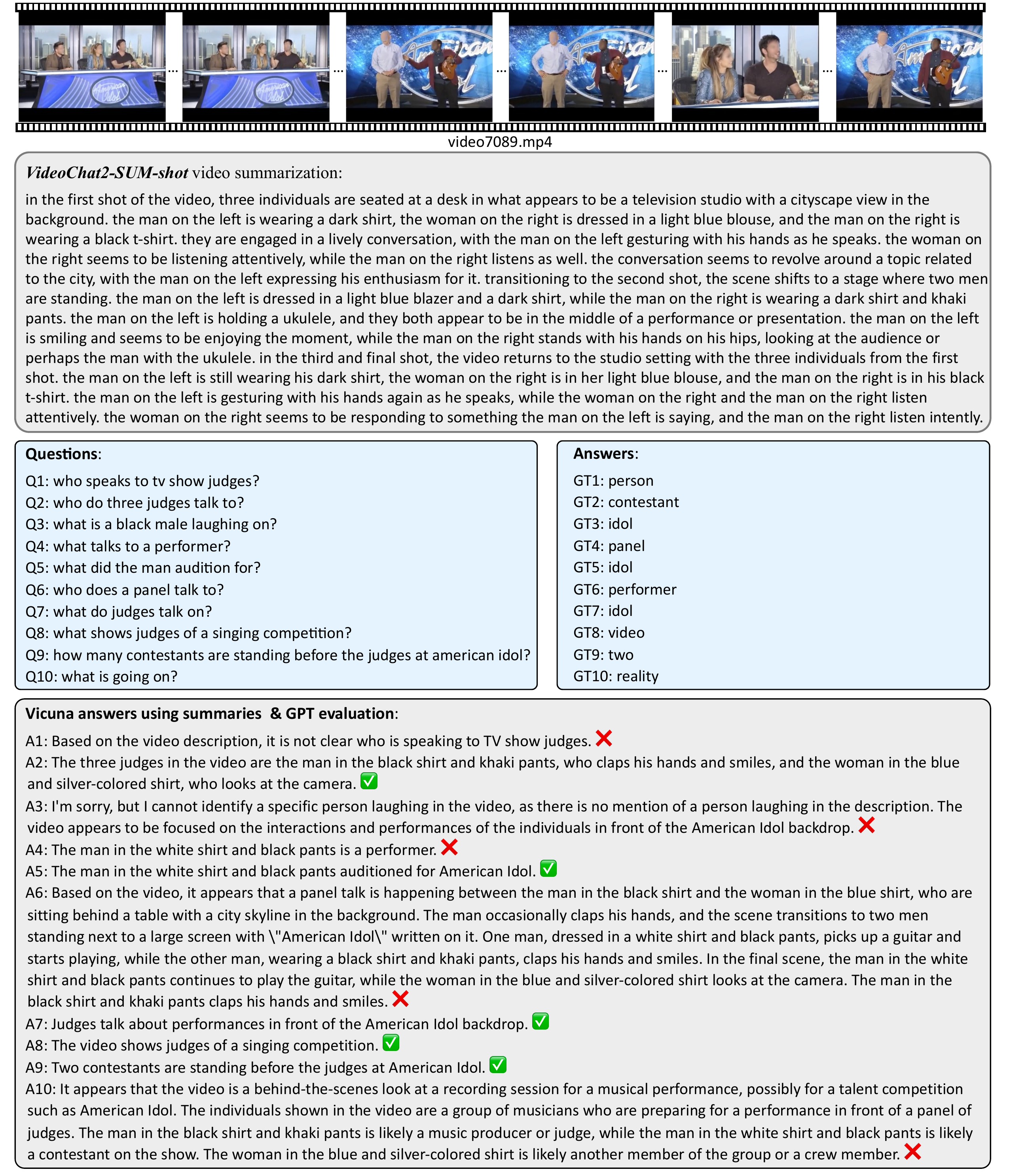}
  \caption{Example for video question-answering using summaries - MSRVTT. We show the zero-shot video summary using our trained VideoChat2-SUM-shot model. The summary can capture most of the video, such as the subject, the action and the transitions. We also show the answers from Vicuna to the first 10 questions and the evaluation results from GPT. Please find more explanations in Sec.~\ref{supp_sec:qa_summary_examples}.
  }
  \label{fig:supp_qa_summary_msrvtt}
  \vspace{-2mm}
\end{figure*}

\vspace{2mm}
\noindent \textbf{MSRVTT-QA:} For instance, in Figure~\ref{fig:supp_qa_summary_msrvtt}, \texttt{video7089} from MSRVTT-QA portrays a TV show outside the domain of \dataset. This genre typically features minimal movement within individual shots, and frequent scene transitions, but a restricted variety of scenes. Yet, our generated summary aptly identifies principal elements such as the judges and contestants, actions like ``engaged in a conversation'' or ``picking up a guitar'', and the setting of an American Idol audition. These details equip the summary to competently address questions from MSRVTT-QA. However, some gaps in detail lead to inaccuracies: of the first 10 questions for \texttt{video7089}, 5 are incorrectly answered due to missing information (\eg, Q1, Q3), incorrect summary content (\eg, Q4), or misalignment with the ground truth (\eg, Q6, Q10).

\begin{figure*}
  \centering
  \includegraphics[width=0.98\linewidth]{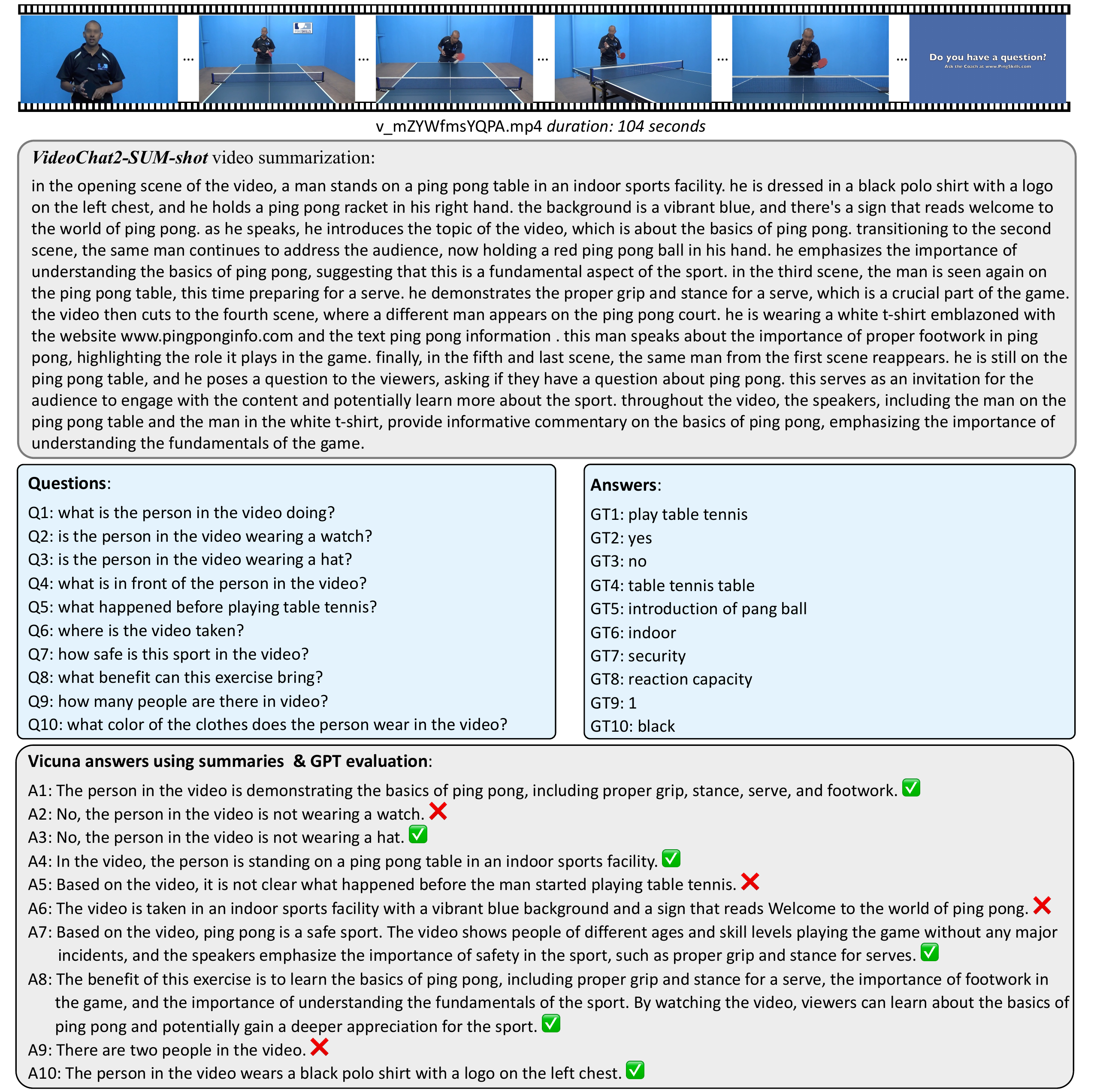}
  \caption{Example for video question-answering using summaries - ActivityNet. We show the zero-shot video summary using our trained VideoChat2-SUM-shot model. The summary can capture most of the video, such as the subject, the action and the transitions. We also show the answers from Vicuna and the evaluation results from GPT. Please find more explanations in Sec.~\ref{supp_sec:qa_summary_examples}.
  }
  \label{fig:supp_qa_summary_anet}
  \vspace{-2mm}
\end{figure*}

\vspace{2mm}
\noindent \textbf{ActivityNet-QA:} In Figure~\ref{fig:supp_qa_summary_anet}, we present the video \texttt{v\_mZYWfmsYQPA} from ActivityNet-QA. The video's duration is 104 seconds, which is significantly longer than the average duration in our \dataset. Our summary effectively identifies important elements such as the main subject's clothing described as ``dressed in a black polo shirt with a logo'', the actions including ``address the audiance'' and ``demonstrate the proper grip'', and the setting, noted as ``a ping pong table in an indoor sports facility''. These comprehensive details enable the Vicuna model to correctly answer most questions from ActivityNet-QA. Despite these insights, some omissions and discrepancies in the summary contribute to inaccuracies in answering. Specifically, four out of ten questions are answered incorrectly due to either a lack of specific details or mismatches with the ground truth, as seen in questions Q2, Q5, Q6 and Q9.

\begin{figure*}
  \centering
  \includegraphics[width=0.98\linewidth]{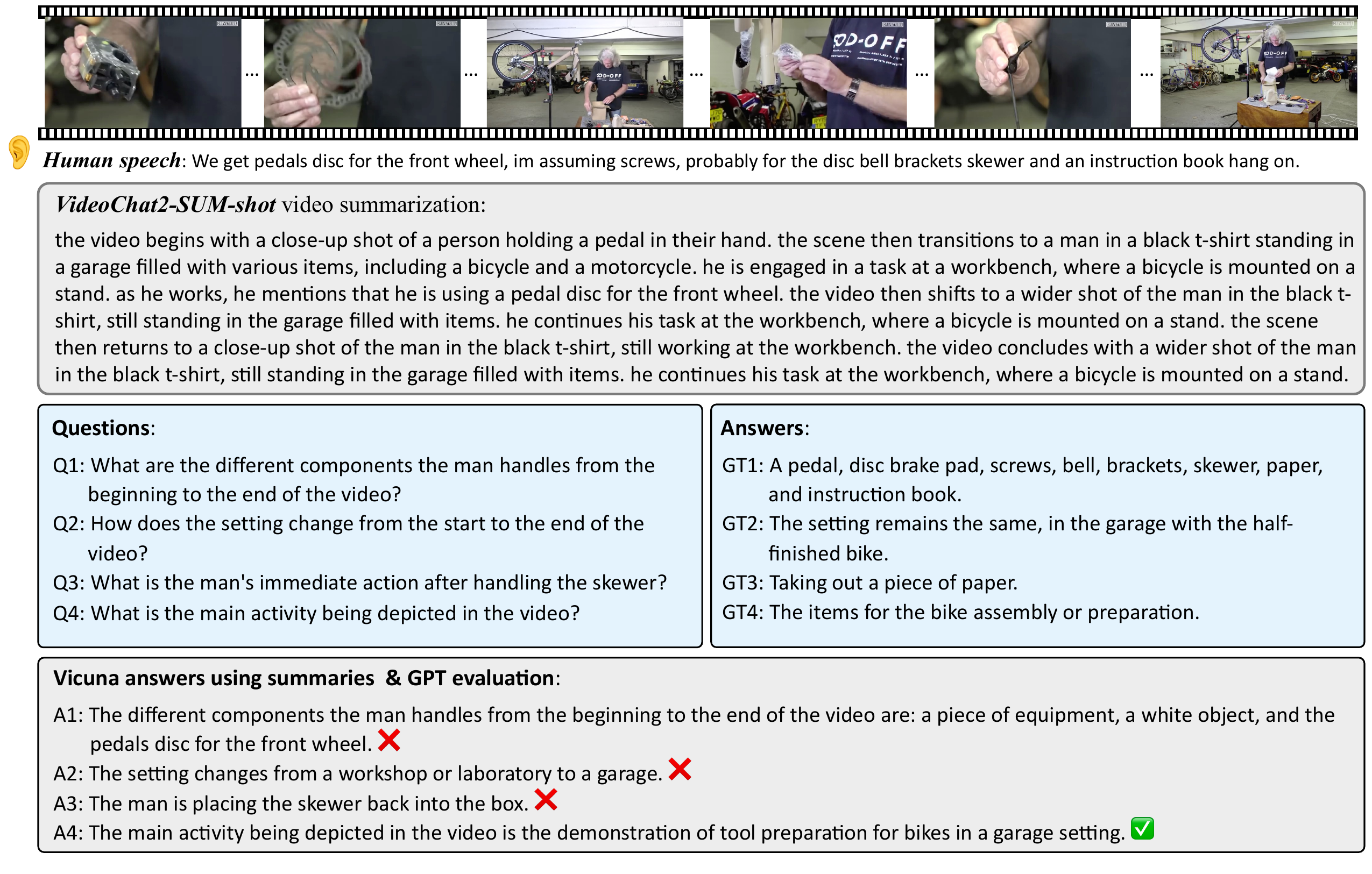}
  \caption{Example for video question-answering using summaries - Shot2Story-QA. We show the in-domain video summary using our trained VideoChat2-SUM-shot model. The summary can capture the shot transitions and major actions. \textcolor{black}{However, as indicated by the QA results, the detailed actions and their order fail to match with the groundtruth, such as whether the immediate action after handing the skewer. Additionally, it fails to provide adequate information regarding the consistent scene between shot transitions, such as ``at a workbench'' and ``in the garage''}. Please find explanations in Sec.~\ref{supp_sec:qa_summary_examples}.
  }
  \label{fig:supp_qa_summary_s2s}
  \vspace{-2mm}
\end{figure*}

\vspace{2mm}
\noindent \textbf{Shot2Story-QA:} 
In Figure~\ref{fig:supp_qa_summary_s2s}, we present a video sample from our Shot2Story-QA dataset. The questions address audio-related (Q1, Q4), holistic understanding (Q2, Q3), and temporal-related (Q4) aspects. With shot structure enabled in VideoChat2-SUM-shot, the generated video summary includes actions and transitions, such as \textit{begins with a close-up shot of a person holding a pedal}, \textit{transitions to a man}, \textit{then shifts to a wider shot}, \textit{returns to a close-up shot of the man}, and \textit{concludes with a wider shot of the man}. However, the generated summary falls short in capturing detailed information about the actions and their orders (e.g., "Taking out a piece of paper" in Q3), and fails to align visual and audio content (e.g., different components in Q1 which have been mentioned in ASR texts). As a result, Vicuna struggles to answer these questions, highlighting the challenges posed by our Shot2Story-QA benchmark. To further improve the model performance, the summarization model could be enhanced with detailed information with denser frames, fine-grained matching of audio and visual cues, and tracking of the same objects across the video. We leave it as future work.


\section{Additional Implementation Details}

\vspace{2mm}
\noindent \textbf{Video-Shot Captioning.} For each video shot, we uniformly sample 4 frames. For testing, a consistent text prompt is used, \ie, ``The audio transcripts are {asr}. Describe this video in detail.''. The maximum number of new tokens generated by the LLM is capped at 180 for both training and inference.

\vspace{2mm}
\noindent \textbf{MiniGPT4-SUM-shot, MiniGPT4-Holistic and VideoChat2-SUM-shot.} In two SUM-shot models, 4 frames per video shot are sampled uniformly. In MiniGPT4-SUM-holistic, 16 frames per video clip are sampled. The rationale behind sampling 16 frames in a holistic approach is based on our dataset's average of 4 shots per video, aligning with the SUM-shot approach of 4 frames per shot.  For both training and inference, the LLM's maximum new token count is set at 600. A consistent text prompt, \ie, ``The audio transcripts are \{asr\}. Describe this video in detail.'', is used during inference.

For video-shot captioning and video summarization tasks, both models are trained on $8\times$2 A100-80G GPUs using Pytorch. The captioning model is trained for 10 epochs, with the best-performing checkpoint on the validation set used for test performance reporting. To prevent overfitting, text prompts are randomly sampled for each sample, as detailed in Sec.~\ref{supp_sec:our_prompt_all}.



\vspace{2mm}
\noindent \textbf{VAST.} We tune the model following the official instructions by inputting our video frames, audios and ASR texts. The optimizing target is the concatenation of our single-shot visual caption and single-shot narration caption. During inference, we input the corresponding modalities, in case of two different model versions we trained (visual + audio, or visual + audio + ASR texts), into the model and use the directly generated captions for evaluation.
The model is trained on 8$\times$4 A100-80G GPUs for 3 epochs, maintaining other hyperparameters at their default values from the original configuration.

\vspace{2mm}
\noindent \textbf{Video-ChatGPT.} Consistent with SUM-holistic, we uniformly sample 16 frames for both training and inference. Our prompt setup excludes ASR for video summarization. The training is conducted over 3 epochs with a learning rate of 2e-6, and we retain other hyperparameters at their default settings as specified in the original repository.

\vspace{2mm}
\noindent \textbf{Video Question-Answering with Summary.} For the MSRVTT-QA~\citep{xu2017msrvttqa} and ActivityNet-QA~\citep{yu2019activitynetqa} datasets, we generate video summaries using our trained SUM-shot model, employing only visual tokens during inference. Upon generating these summaries, we integrate them with individual questions from their corresponding videos into the prompt format displayed in Figure~\ref{fig:supp_qa_prompt}. This integrated content is then processed through Vicuna~\citep{vicuna2023} to obtain answers. The evaluation of these results is carried out following the methodology outlined in \citep{maaz2023videochatgpt}.

\section{Broader Impact}
\noindent \textbf{Data Limitations and Ethical Considerations.} We provide cropped multi-shot videos instead of the original videos. Users can also turn to download these from original sources. Given HD-VILA-100M~\citep{xue2022hdvila}'s long-standing public availability, we assess a low risk of the currently available videos being removed in the near future. Additionally, our meticulous manual annotation process is designed to avoid any ethical or legal violations. Specifically, our videos don't have personally identifiable information or offensive content, which is ensured by the manual annotation process. The authors will take the responsibility of long-term maintenance. 

\noindent \textbf{Human Rights in Annotation Process.} We have conscientiously structured the annotation process to ensure fair workloads and equitable compensation for annotators, upholding human rights standards.

\noindent \textbf{Scope of Conclusions.} It is important to recognize that experiments and data, including ours, might only represent a subset of universal realities. Nevertheless, given the wide range of categories covered in our videos, we believe our conclusions offer a robust understanding applicable to various multi-shot video scenarios and durations. These findings, while specific to our dataset, provide significant insight into the broader field of video analysis.

\noindent \textbf{Usage of Language Models.} Our use of the LLaMA model~\citep{touvron2023llama} from Meta is authorized for research purposes. Those intending to use our model post-release should ensure that they have the necessary permissions and adhere to usage restrictions. We express deep respect for the work of developers and contributors, recognizing their integral role in advancing language modelling and multi-modal learning.

\noindent \textbf{Future Research and Development.}
We release both our code and dataset. This is intended to encourage further research and enable others to build upon our work. Although our current experiments require up to 8$\times$2 A100-80G GPUs, we are aware this may be a limitation.
Consequently, we plan to focus future efforts on adapting these experiments to be compatible with a single node of 8 A100 GPUs. It's important to note that fitting the experiments within an 8 GPU framework is not the primary focus of this paper, but we consider it a crucial step towards making our research more accessible and inclusive for a wider array of research groups.



%% file: main.bbl
\begin{thebibliography}{38}
\providecommand{\natexlab}[1]{#1}
\providecommand{\url}[1]{\texttt{#1}}
\expandafter\ifx\csname urlstyle\endcsname\relax
  \providecommand{\doi}[1]{doi: #1}\else
  \providecommand{\doi}{doi: \begingroup \urlstyle{rm}\Url}\fi

\bibitem[Achiam et~al.(2023)Achiam, Adler, Agarwal, Ahmad, Akkaya, Aleman, Almeida, Altenschmidt, Altman, Anadkat, et~al.]{achiam2023gpt}
Josh Achiam, Steven Adler, Sandhini Agarwal, Lama Ahmad, Ilge Akkaya, Florencia~Leoni Aleman, Diogo Almeida, Janko Altenschmidt, Sam Altman, Shyamal Anadkat, et~al.
\newblock Gpt-4 technical report.
\newblock \emph{arXiv preprint arXiv:2303.08774}, 2023.

\bibitem[Anne~Hendricks et~al.(2017)Anne~Hendricks, Wang, Shechtman, Sivic, Darrell, and Russell]{anne2017localizing}
Lisa Anne~Hendricks, Oliver Wang, Eli Shechtman, Josef Sivic, Trevor Darrell, and Bryan Russell.
\newblock Localizing moments in video with natural language.
\newblock In \emph{ICCV}, pp.\  5803--5812, 2017.

\bibitem[Chen et~al.(2024)Chen, Li, Wang, Zhao, Sun, Zhu, and Liu]{chen2024vast}
Sihan Chen, Handong Li, Qunbo Wang, Zijia Zhao, Mingzhen Sun, Xinxin Zhu, and Jing Liu.
\newblock Vast: A vision-audio-subtitle-text omni-modality foundation model and dataset.
\newblock \emph{Advances in Neural Information Processing Systems}, 36, 2024.

\bibitem[Chiang et~al.(2023)Chiang, Li, Lin, Sheng, Wu, Zhang, Zheng, Zhuang, Zhuang, Gonzalez, Stoica, and Xing]{vicuna2023}
Wei-Lin Chiang, Zhuohan Li, Zi~Lin, Ying Sheng, Zhanghao Wu, Hao Zhang, Lianmin Zheng, Siyuan Zhuang, Yonghao Zhuang, Joseph~E. Gonzalez, Ion Stoica, and Eric~P. Xing.
\newblock Vicuna: An open-source chatbot impressing gpt-4 with 90\%* chatgpt quality, March 2023.
\newblock URL \url{https://lmsys.org/blog/2023-03-30-vicuna/}.

\bibitem[Cutting et~al.(2011)Cutting, Brunick, DeLong, Iricinschi, and Candan]{cutting2011shot}
James~E Cutting, Kaitlin~L Brunick, Jordan~E DeLong, Catalina Iricinschi, and Ayse Candan.
\newblock Quicker, faster, darker: Changes in hollywood film over 75 years.
\newblock \emph{i-Perception}, 2\penalty0 (6):\penalty0 569--576, 2011.

\bibitem[Denkowski \& Lavie(2014)Denkowski and Lavie]{denkowski2014meteor}
Michael Denkowski and Alon Lavie.
\newblock Meteor universal: Language specific translation evaluation for any target language.
\newblock In \emph{Proceedings of the ninth workshop on statistical machine translation}, pp.\  376--380, 2014.

\bibitem[Fang et~al.(2022)Fang, Wang, Xie, Sun, Wu, Wang, Huang, Wang, and Cao]{EVA}
Yuxin Fang, Wen Wang, Binhui Xie, Quan Sun, Ledell Wu, Xinggang Wang, Tiejun Huang, Xinlong Wang, and Yue Cao.
\newblock Eva: Exploring the limits of masked visual representation learning at scale.
\newblock \emph{arXiv preprint arXiv:2211.07636}, 2022.

\bibitem[Gemmeke et~al.(2017)Gemmeke, Ellis, Freedman, Jansen, Lawrence, Moore, Plakal, and Ritter]{gemmeke2017audioset}
Jort~F Gemmeke, Daniel~PW Ellis, Dylan Freedman, Aren Jansen, Wade Lawrence, R~Channing Moore, Manoj Plakal, and Marvin Ritter.
\newblock Audio set: An ontology and human-labeled dataset for audio events.
\newblock In \emph{2017 IEEE international conference on acoustics, speech and signal processing (ICASSP)}, pp.\  776--780. IEEE, 2017.

\bibitem[Grauman et~al.(2022)Grauman, Westbury, Byrne, Chavis, Furnari, Girdhar, Hamburger, Jiang, Liu, Liu, et~al.]{grauman2022ego4d}
Kristen Grauman, Andrew Westbury, Eugene Byrne, Zachary Chavis, Antonino Furnari, Rohit Girdhar, Jackson Hamburger, Hao Jiang, Miao Liu, Xingyu Liu, et~al.
\newblock Ego4d: Around the world in 3,000 hours of egocentric video.
\newblock In \emph{CVPR}, 2022.

\bibitem[Guo et~al.(2023)Guo, Li, Li, Tiong, Li, Tao, and Hoi]{guo2023imagestextualprompt}
Jiaxian Guo, Junnan Li, Dongxu Li, Anthony Meng~Huat Tiong, Boyang Li, Dacheng Tao, and Steven Hoi.
\newblock From images to textual prompts: Zero-shot visual question answering with frozen large language models.
\newblock In \emph{Proceedings of the IEEE/CVF Conference on Computer Vision and Pattern Recognition}, pp.\  10867--10877, 2023.

\bibitem[Hu et~al.(2021)Hu, Shen, Wallis, Allen-Zhu, Li, Wang, Wang, and Chen]{hu2021lora}
Edward~J Hu, Yelong Shen, Phillip Wallis, Zeyuan Allen-Zhu, Yuanzhi Li, Shean Wang, Lu~Wang, and Weizhu Chen.
\newblock Lora: Low-rank adaptation of large language models.
\newblock \emph{arXiv preprint arXiv:2106.09685}, 2021.

\bibitem[Kim et~al.(2021)Kim, Heo, Choe, Chung, Kwon, Lee, Kwon, and Chung]{kim2021speakerdet}
You~Jin Kim, Hee-Soo Heo, Soyeon Choe, Soo-Whan Chung, Yoohwan Kwon, Bong-Jin Lee, Youngki Kwon, and Joon~Son Chung.
\newblock Look who's talking: Active speaker detection in the wild.
\newblock \emph{Interspeech}, 2021.

\bibitem[Krishna et~al.(2017)Krishna, Hata, Ren, Fei-Fei, and Carlos~Niebles]{krishna2017activitynet}
Ranjay Krishna, Kenji Hata, Frederic Ren, Li~Fei-Fei, and Juan Carlos~Niebles.
\newblock Dense-captioning events in videos.
\newblock In \emph{Proceedings of the IEEE international conference on computer vision}, pp.\  706--715, 2017.

\bibitem[Li et~al.(2019)Li, Wong, Zhao, and Kankanhalli]{li2019video}
Junnan Li, Yongkang Wong, Qi~Zhao, and Mohan~S Kankanhalli.
\newblock Video storytelling: Textual summaries for events.
\newblock \emph{IEEE Transactions on Multimedia}, 22\penalty0 (2):\penalty0 554--565, 2019.

\bibitem[Li et~al.(2023{\natexlab{a}})Li, Li, Savarese, and Hoi]{li2023blip2}
Junnan Li, Dongxu Li, Silvio Savarese, and Steven Hoi.
\newblock Blip-2: Bootstrapping language-image pre-training with frozen image encoders and large language models.
\newblock \emph{arXiv preprint arXiv:2301.12597}, 2023{\natexlab{a}}.

\bibitem[Li et~al.(2023{\natexlab{b}})Li, He, Wang, Li, Wang, Luo, Wang, Wang, and Qiao]{li2023videochat}
KunChang Li, Yinan He, Yi~Wang, Yizhuo Li, Wenhai Wang, Ping Luo, Yali Wang, Limin Wang, and Yu~Qiao.
\newblock Videochat: Chat-centric video understanding, 2023{\natexlab{b}}.

\bibitem[Li et~al.(2023{\natexlab{c}})Li, Wang, He, Li, Wang, Liu, Wang, Xu, Chen, Luo, et~al.]{li2023mvbench}
Kunchang Li, Yali Wang, Yinan He, Yizhuo Li, Yi~Wang, Yi~Liu, Zun Wang, Jilan Xu, Guo Chen, Ping Luo, et~al.
\newblock Mvbench: A comprehensive multi-modal video understanding benchmark.
\newblock \emph{arXiv preprint arXiv:2311.17005}, 2023{\natexlab{c}}.

\bibitem[Li et~al.(2023{\natexlab{d}})Li, Wang, Li, Wang, He, Wang, and Qiao]{li2023unmasked}
Kunchang Li, Yali Wang, Yizhuo Li, Yi~Wang, Yinan He, Limin Wang, and Yu~Qiao.
\newblock Unmasked teacher: Towards training-efficient video foundation models.
\newblock \emph{arXiv preprint arXiv:2303.16058}, 2023{\natexlab{d}}.

\bibitem[Li et~al.(2023{\natexlab{e}})Li, Wang, and Jia]{li2023llama}
Yanwei Li, Chengyao Wang, and Jiaya Jia.
\newblock Llama-vid: An image is worth 2 tokens in large language models.
\newblock \emph{arXiv preprint arXiv:2311.17043}, 2023{\natexlab{e}}.

\bibitem[Lin et~al.(2023)Lin, Zhu, Ye, Ning, Jin, and Yuan]{lin2023video}
Bin Lin, Bin Zhu, Yang Ye, Munan Ning, Peng Jin, and Li~Yuan.
\newblock Video-llava: Learning united visual representation by alignment before projection.
\newblock \emph{arXiv preprint arXiv:2311.10122}, 2023.

\bibitem[Lin(2004)]{lin2004rouge}
Chin-Yew Lin.
\newblock Rouge: A package for automatic evaluation of summaries.
\newblock In \emph{Text summarization branches out}, pp.\  74--81, 2004.

\bibitem[Loshchilov \& Hutter(2017)Loshchilov and Hutter]{loshchilov2017adamw}
Ilya Loshchilov and Frank Hutter.
\newblock Decoupled weight decay regularization.
\newblock \emph{arXiv preprint arXiv:1711.05101}, 2017.

\bibitem[Maaz et~al.(2023)Maaz, Rasheed, Khan, and Khan]{maaz2023videochatgpt}
Muhammad Maaz, Hanoona Rasheed, Salman Khan, and Fahad~Shahbaz Khan.
\newblock Video-chatgpt: Towards detailed video understanding via large vision and language models.
\newblock \emph{arXiv preprint arXiv:2306.05424}, 2023.

\bibitem[OpenAI()]{chatgpt}
OpenAI.
\newblock Gpt-4.
\newblock URL \url{https://chat.openai.com/?model=gpt-4}.

\bibitem[Papineni et~al.(2002)Papineni, Roukos, Ward, and Zhu]{papineni2002bleu}
Kishore Papineni, Salim Roukos, Todd Ward, and Wei-Jing Zhu.
\newblock Bleu: a method for automatic evaluation of machine translation.
\newblock In \emph{Proceedings of the 40th annual meeting of the Association for Computational Linguistics}, pp.\  311--318, 2002.

\bibitem[Radford et~al.(2021)Radford, Kim, Hallacy, Ramesh, Goh, Agarwal, Sastry, Askell, Mishkin, Clark, et~al.]{radford2021learning}
Alec Radford, Jong~Wook Kim, Chris Hallacy, Aditya Ramesh, Gabriel Goh, Sandhini Agarwal, Girish Sastry, Amanda Askell, Pamela Mishkin, Jack Clark, et~al.
\newblock Learning transferable visual models from natural language supervision.
\newblock In \emph{International conference on machine learning}, pp.\  8748--8763. PMLR, 2021.

\bibitem[Song et~al.(2023)Song, Chai, Wang, Zhang, Zhou, Wu, Guo, Ye, Lu, Hwang, et~al.]{song2023moviechat}
Enxin Song, Wenhao Chai, Guanhong Wang, Yucheng Zhang, Haoyang Zhou, Feiyang Wu, Xun Guo, Tian Ye, Yan Lu, Jenq-Neng Hwang, et~al.
\newblock Moviechat: From dense token to sparse memory for long video understanding.
\newblock \emph{arXiv preprint arXiv:2307.16449}, 2023.

\bibitem[Sou{\v{c}}ek \& Loko{\v{c}}(2020)Sou{\v{c}}ek and Loko{\v{c}}]{souvcek2020transnet}
Tom{\'a}{\v{s}} Sou{\v{c}}ek and Jakub Loko{\v{c}}.
\newblock Transnet v2: An effective deep network architecture for fast shot transition detection.
\newblock \emph{arXiv preprint arXiv:2008.04838}, 2020.

\bibitem[Touvron et~al.(2023)Touvron, Lavril, Izacard, Martinet, Lachaux, Lacroix, Rozi{\`e}re, Goyal, Hambro, Azhar, et~al.]{touvron2023llama}
Hugo Touvron, Thibaut Lavril, Gautier Izacard, Xavier Martinet, Marie-Anne Lachaux, Timoth{\'e}e Lacroix, Baptiste Rozi{\`e}re, Naman Goyal, Eric Hambro, Faisal Azhar, et~al.
\newblock Llama: Open and efficient foundation language models.
\newblock \emph{arXiv preprint arXiv:2302.13971}, 2023.

\bibitem[Vedantam et~al.(2015)Vedantam, Lawrence~Zitnick, and Parikh]{vedantam2015cider}
Ramakrishna Vedantam, C~Lawrence~Zitnick, and Devi Parikh.
\newblock Cider: Consensus-based image description evaluation.
\newblock In \emph{Proceedings of the IEEE conference on computer vision and pattern recognition}, pp.\  4566--4575, 2015.

\bibitem[Xu et~al.(2017)Xu, Zhao, Xiao, Wu, Zhang, He, and Zhuang]{xu2017msrvttqa}
Dejing Xu, Zhou Zhao, Jun Xiao, Fei Wu, Hanwang Zhang, Xiangnan He, and Yueting Zhuang.
\newblock Video question answering via gradually refined attention over appearance and motion.
\newblock In \emph{ACM Multimedia}, 2017.

\bibitem[Xu et~al.(2016)Xu, Mei, Yao, and Rui]{xu2016msr-vtt}
Jun Xu, Tao Mei, Ting Yao, and Yong Rui.
\newblock Msr-vtt: A large video description dataset for bridging video and language.
\newblock CVPR, June 2016.

\bibitem[Xue et~al.(2022)Xue, Hang, Zeng, Sun, Liu, Yang, Fu, and Guo]{xue2022hdvila}
Hongwei Xue, Tiankai Hang, Yanhong Zeng, Yuchong Sun, Bei Liu, Huan Yang, Jianlong Fu, and Baining Guo.
\newblock Advancing high-resolution video-language representation with large-scale video transcriptions.
\newblock In \emph{CVPR}, pp.\  5036--5045, 2022.

\bibitem[Yu et~al.(2019)Yu, Xu, Yu, Yu, Zhao, Zhuang, and Tao]{yu2019activitynetqa}
Zhou Yu, Dejing Xu, Jun Yu, Ting Yu, Zhou Zhao, Yueting Zhuang, and Dacheng Tao.
\newblock Activitynet-qa: A dataset for understanding complex web videos via question answering.
\newblock In \emph{Proceedings of the AAAI Conference on Artificial Intelligence}, 2019.

\bibitem[Zhang et~al.(2023)Zhang, Lu, Islam, Wang, Yu, Bansal, and Bertasius]{zhang2023simple}
Ce~Zhang, Taixi Lu, Md~Mohaiminul Islam, Ziyang Wang, Shoubin Yu, Mohit Bansal, and Gedas Bertasius.
\newblock A simple llm framework for long-range video question-answering.
\newblock \emph{arXiv preprint arXiv:2312.17235}, 2023.

\bibitem[Zhou et~al.(2018)Zhou, Xu, and Corso]{zhou2018youcook2}
Luowei Zhou, Chenliang Xu, and Jason Corso.
\newblock Towards automatic learning of procedures from web instructional videos.
\newblock In \emph{AAAI}, volume~32, 2018.

\bibitem[Zhou et~al.(2019)Zhou, Kalantidis, Chen, Corso, and Rohrbach]{zhou2019grounded}
Luowei Zhou, Yannis Kalantidis, Xinlei Chen, Jason~J Corso, and Marcus Rohrbach.
\newblock Grounded video description.
\newblock In \emph{CVPR}, pp.\  6578--6587, 2019.

\bibitem[Zhu et~al.(2023)Zhu, Chen, Shen, Li, and Elhoseiny]{zhu2023minigpt}
Deyao Zhu, Jun Chen, Xiaoqian Shen, Xiang Li, and Mohamed Elhoseiny.
\newblock Minigpt-4: Enhancing vision-language understanding with advanced large language models.
\newblock \emph{arXiv preprint arXiv:2304.10592}, 2023.

\end{thebibliography}
